\documentclass[preprint,11pt,twocolumn]{elsarticle}

\usepackage[margin={1.5cm,1.5cm}]{geometry}
\usepackage{amsmath,amssymb,amsfonts}
\usepackage{hyperref}
\usepackage{array,multirow,mathtools,afterpage }
\usepackage{longtable}
\usepackage{latexsym,color}
\usepackage{graphicx, subfig}
\usepackage[linesnumbered,ruled,vlined]{algorithm2e}
\usepackage{algpseudocode}
\usepackage{array}
\usepackage{ragged2e}
\newcolumntype{P}[1]{>{\RaggedRight\hspace{0pt}}p{#1}}

\journal{Knowledge-Based Systems}

\begin{document}

\twocolumn[
	\begin{@twocolumnfalse}
\begin{frontmatter}

\title{Story Disambiguation: Tracking Evolving News Stories across News and Social Streams}
%\tnotetext[mytitlenote]{Fully documented templates are available in the elsarticle package on \href{http://www.ctan.org/tex-archive/macros/latex/contrib/elsarticle}{CTAN}.}

%% Group authors per affiliation:
\author{Bichen Shi, Thanh-Binh Le, Neil Hurley and Georgiana Ifrim \\
{\small \{bichen.shi,thanh.binh,neil.hurley,georgiana.ifrim\}@insight-centre.org }}
\address{Insight Centre for Data Analytics, University College Dublin, Ireland}
%\ead[]{{bichen.shi,thanh.binh,georgiana.ifrim,neil.hurley}@insight-centre.org}
%\fntext[myfootnote]{Since 1880.}

%% or include affiliations in footnotes:
%\author[mymainaddress,mysecondaryaddress]{Elsevier Inc}
%\ead[url]{www.elsevier.com}

%\author[mysecondaryaddress]{Global Customer Service\corref{mycorrespondingauthor}}
%\cortext[mycorrespondingauthor]{Corresponding author}
%\ead{support@elsevier.com}

%\address[mymainaddress]{1600 John F Kennedy Boulevard, Philadelphia}
%\address[mysecondaryaddress]{360 Park Avenue South, New York}

\begin{abstract}
%What is the problem?
%Why is it important?
%Why is it challenging?
Following a particular news story online is an important but difficult task, as the relevant information is often scattered across different domains/sources
(e.g., news articles, blogs, comments, tweets), presented in various formats and language styles, and may overlap with thousands of other stories. 
%What is the state-of-the-art?
%What do we propose?
In this work we join the areas of topic tracking and entity disambiguation, and propose a framework named Story Disambiguation -  
a cross-domain story tracking approach that builds on real-time entity disambiguation and a learning-to-rank framework 
to represent and update the rich semantic structure of news stories. Given a target news story, specified by a seed set of documents, 
the goal is to effectively select new story-relevant documents from an incoming document stream.  
%(e.g., news articles, tweets).
We represent stories as entity graphs and we model the story tracking problem as a learning-to-rank task. 
This enables us to track content with high accuracy, from multiple domains, in real-time.
% where new documents arrive as a stream, and 
%the task is to 
We study a range of text, entity and graph based features to understand which type of features are most effective for representing stories.
We further propose new semi-supervised learning techniques to automatically update the story representation over time.
%We employ the entity disambiguation to increase the overall accuracy of the approach.
%employ entity graphs for news story representation, together with entity disambiguation, 
%learning-to-rank framework, and semi-supervised learning techniques, to allow tracking content with high accuracy from multiple sources in real-time. 
%What have we learned form this study?
Our empirical study shows that we outperform the accuracy of state-of-the-art methods for tracking mixed-domain document streams, 
while requiring fewer labeled data to seed the tracked stories.
This is particularly the case for local news stories that are easily over shadowed by other trending stories, 
and for complex news stories with ambiguous content in noisy stream environments.
%where a rich amount of the story details are reported in the news.
\end{abstract}


\end{frontmatter}
\end{@twocolumnfalse}]
%\linenumbers

\section{Introduction}
\label{sec:intro}
%!TEX root = main.tex

Following a particular news story online is an important but difficult task, as the relevant information is often scattered across different domains/sources (e.g., news articles, blogs, comments, posts, and tweets), embedded in various formats and language styles, and flooded by thousands of other stories. 
The well-studied research area of Topic Detection and Tracking (TDT) focuses on detecting new stories and tracking a particular story from a collection of long and informative documents written in formal language (e.g., news articles).
With the rise of social media, more often, the relevant information is embedded in short, fragmented, noisy and ambiguous content, making the story tracking task very challenging.
%This is particularly the case if we want to use a unified approach for tracking content from both news and social streams, where the language styles, content length and the level of noise are significantly different.

\begin{figure}
    \centering
    \includegraphics[width=0.8\linewidth, ]{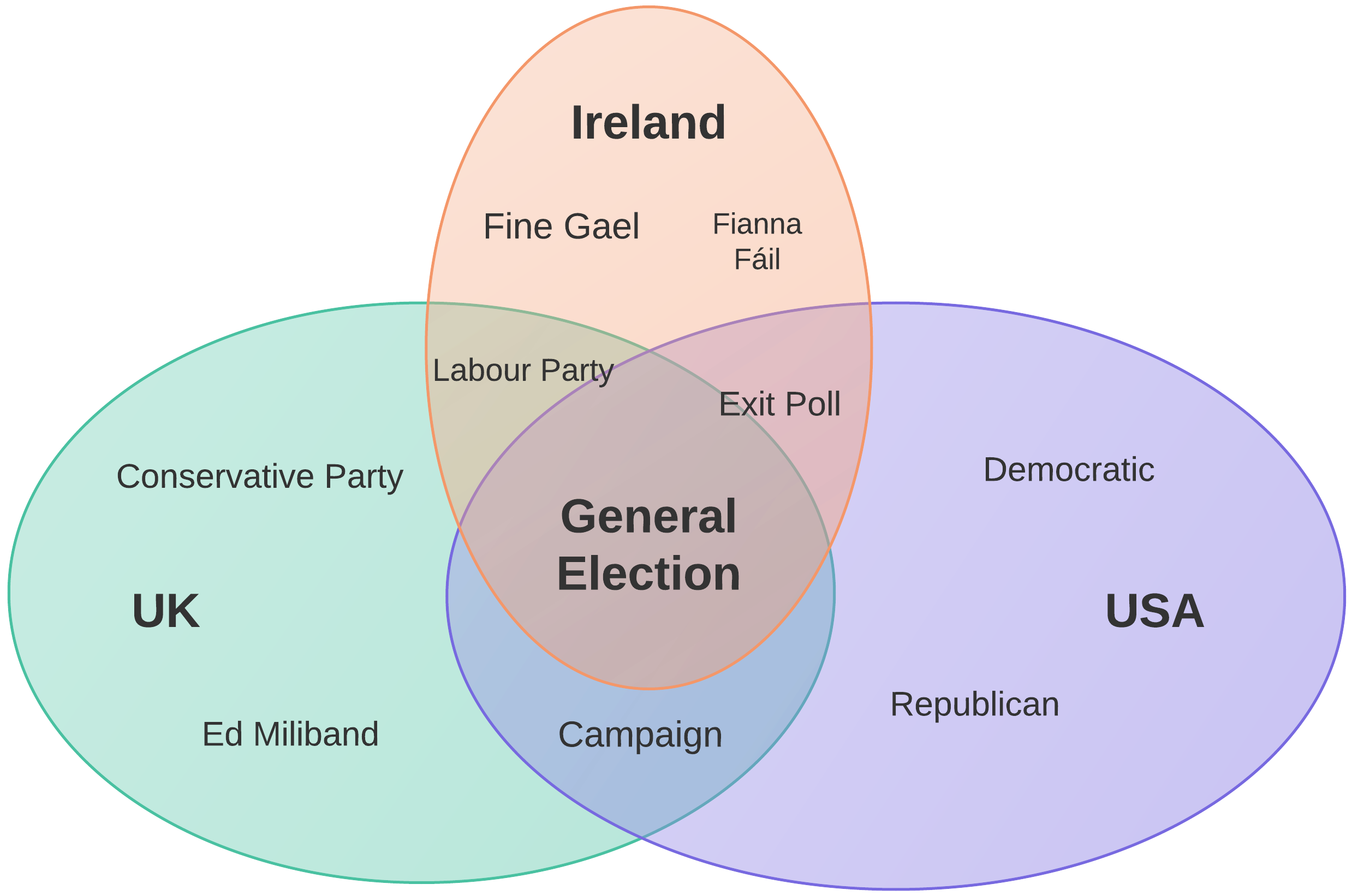}
    \caption{Phrases overlapping between the three general election stories: Irish, UK and US general elections, 2016.}
    \label{fig:ge16-irish-uk-us}
\end{figure}

When tracking a story on news data, some TDT methods take a binary classification approach, by considering the target story as one class, and irrelevant documents as another \cite{allan1998topic,schultz1999topic}.
A single classifier is typically trained off-line on a relatively small batch of manually labeled data.
Probabilistic topic modeling \cite{brants2009systems,wu2011domain} has also been successfully applied to predict the underlying structure of text data.
The model in this case learns the distribution of underlying topics (i.e., stories) of documents.
However, these techniques need a large amount of positive and negative training examples for each story, and are inflexible when dealing with story drift over time.
Limited by the type of features typically used (e.g., bag-of-words), such approaches can only work well on one type of content, namely the one used for training (e.g., news articles), 
and cannot be directly applied to other types of content (e.g., tweets).  
On the other hand, when tracking a story on social streams, many existing approaches use key-phrases to define stories and monitor the abrupt spikes in the appearance of the story key-phrases \cite{adar2004implicit,gruhl2004information,kumar2004structure}.
The different choice of modeling when handling news and social data is due to the distinct nature of these two types of data.
Because social media data tends to be large scale and have a short and fragmented format, the tracking model must be efficient and only use light-weight features. 
However, due to word ambiguity and the short content, tracking text phrases alone, without identifying their true meaning, may result in retrieving noisy content.
To clarify these points, we show an example in Figure \ref{fig:ge16-irish-uk-us}: when tracking the story "Irish General Election 2016", across news and social media, 
one could easily mix up the content of the Irish, US, and UK general elections published in the same period, in particular, when the nationality is not highlighted in the text, which is often the case.
Most state-of-the-art (SOTA) story tracking approaches focus on only one type of data, and very few prior works propose unified approaches that can track stories on different domains simultaneously (i.e., both news and social streams).
However, there is enormous potential in efficiently tracking content from multiple data streams, as this can enable the flow of information between the content of different domains, and thus, 
present a more complete picture of the story being tracked.
%Thus, the ability of correctly understand the underlying meaning of the text phrases is critical when tracking story online, and we take advantage of our work in Chapter \ref{chap:disambiguation}
%The ability to disambiguating the entities appeared in the evolving news stories is important in story tracking, to better define stories and to retrieve content across sources with high accuracy.
%An automatic approach to disambiguating the entities appearing in evolving news stories is needed, to better define and understand stories, and to retrieve content across sources with high accuracy.
%This involves efforts in two directions: topic tracking and entity disambiguation. 
%The important entities that act as the skeleton of the target story should be identified, weighted, and updated while the story develops over time. 
%The ambiguity of these entities should be resolved by mapping them to the correct meaning in a structured knowledge base (e.g., Wikipedia).

In this paper, we study the problem of story tracking across news and social media. 
Given a target story, specified by a set of seed documents, we aim to effectively classify streaming documents (e.g., news articles, blogs, comments, posts, and tweets) as to whether or not they are relevant to the target news story.
%The target story in our setting is defined by a seed set of news articles and tweets.
%We join the area of topic tracking and entity disambiguation, and propose a framework named Story Disambiguation: a cross-domain story tracking approach that builds on real-time entity disambiguation 
%to represent stories using their rich semantic structure. 
We join the areas of topic tracking and entity disambiguation, and propose a framework named Story Disambiguation -  
a cross-domain story tracking approach that builds on real-time entity disambiguation and a learning-to-rank (L2R) framework to represent 
and update the rich semantic structure of news stories.
We employ entity disambiguation to correctly interpret the underlying meaning of the text, and to easily connect content from different sources, according to their topics.
This allows us to overcome the high level of noise in social media data and to significantly increase the story tracking accuracy on short and fragmented content.
This is especially important when tracking local news stories that are less popular and less visible across the world, 
where the amount of relevant social posts is small, and easily flooded by other trending stories with similar entities 
(e.g., ''Irish water charges'' and ''mortgage crisis in Ireland''\footnote{We have created a page to track some of the Irish local stories: \url{http://insight4news.ucd.ie/issues_ge16/}} are big news stories within Ireland, but not popular across the world.).

To enable our approach to track content from different sources and in distinct language styles, we represent each tracked story in a high-level entity graph, where the entities are the disambiguated text phrases.
Most often, the story is centered around a few key people, organizations, locations, etc., and thus the entity graph is a very effective way of representing a story.
In contrast to TDT approaches that use bag-of-words features, our approach extracts features from both the entity ranking in the graph and the documents of the story.
Thus, in our tracking model, we are robust to the impact of language style (e.g., the length of the documents, vocabulary).
We also exploit the importance of entities (e.g., people, locations, organizations, event names) in the text, which is far more important in defining and tracking stories, than the text similarity.

Most prior approaches train a binary classifier per tracked story and thus need a large amount of labeled data to train each classifier.
Reducing the amount of labeled data affects the tracking accuracy significantly.
We take a different view and model the task of story tracking as an L2R problem by classifying story-document (story-doc) pairs as relevant or irrelevant.
%From our analysis, the model of what makes a document relevant to an story doesn't change over time, and 
Our previous work on social tag recommendation \cite{shi2017hashtagger} shows that an L2R relevance classifier is stable in a highly dynamic environment and does not require retraining to maintain high accuracy. 
In the story tracking problem, the story profile changes over time, the incoming documents are always new, and the target stories can be quite different to each other.
However, what makes a document relevant to a target story does not change.
For example, the central entities in a document and the central entities in a story play a key role in establishing relevance.
An L2R model trained on labeled story-doc pairs can learn to map the features describing any new story-doc pair to a relevance score (since it learns what makes an object, i.e., a doc, relevant to another object, i.e., a story).
This means that we can train a model on a set of past stories and apply it to track new stories, without having to re-train the model.
%to map features describing story-doc pairs to a relevance score
%A pre-trained L2R model that captures the relevance of labeled story-doc pairs, does not need to be retrained for each story, 
%and thus significantly reduces the amount of required labeled data per story.
In terms of the concept drift, many SOTA techniques use active learning and semi-supervised learning techniques to constantly update their training data and classifiers for the target story.
We propose a new semi-supervised learning approach to automatically update the story entity graph and the feature vectors, to reflect the story drift over time.
Unlike classic semi-supervised learning where the training data and the classifiers are updated, our pre-trained relevance classifier remains the same, 
but the features that are describing each example are updated.
This method also helps to reduce the required labeled data for defining the target story.
%The experimental results show that we outperform the accuracy of the state-of-the-art approaches using much fewer labeled data, 
%especially when tracking local news stories, which are easily overshadowed by other trending stories, and complex news stories, where a rich amount of the story details are reported in the news.

We summarize our contributions as follows:

\begin{itemize} 

\item We investigate a unified approach for tracking a specific news story across news and social media, with \textit{high precision}, and \textit{high recall}. 

\item We represent the news story in a high-level entity graph, and continuously update it using semi-supervised learning techniques. The entity graph representation enables us to track content from sources with significantly different language styles, such as news media and social streams.

\item We leverage the relevance of the story-document pairs, and model the story tracking problem as a learning-to-rank task, and thus greatly reduce the need for labeled data for specifying the target story. 

\item We present experiments with different story tracking approaches on news and social media data, and show that our approach outperforms the SOTA in F1, 
especially when tracking local news stories and complex news stories.
\end{itemize}

The rest of this paper is organized as follows. 
Section \ref{sec:rel_story} discusses related work.
In Section \ref{sec:story-method} we present our story tracking  model.
%We describe our methodology and the proposed features in Section \ref{sec:l2r4hashrec}. 
In Section \ref{sec:story-evaluation} we present the evaluation setting and experimental results for our approach as well as a comparison to the state-of-the-art. 
%In Section \ref{sec:applic} we discuss a real-world application of our recommendations to the task of automated story tracking.
%We conclude in 
%experiment evaluating the impact of hashtag recommendation on news engagement, and present an exciting new application, social indexing, a form of real-time
%crowdsourced tagging of news. 
We conclude in Section \ref{sec:story_conclusion}.

%\section{Introduction}\label{sec:intro}
% give use case
% challenge

\section{Related Work}
\label{sec:rel_story}
%!TEX root = main.tex

Topic Detection and Tracking (TDT) has been researched extensively on each of the domains of news articles \cite{alsumait2008line,leek2002probabilistic,wayne2000multilingual,makkonen2004simple,lavrenko2002relevance}, tweets \cite{petrovic2010streaming,lin2011smoothing,magdy2014adaptive} and blogs \cite{tang2011wikipedia}.
However, very few prior works provide a unified solution for topic tracking across domains.
In the following, we discuss related work for topic tracking in different domains.

\subsection{Topic Detection and Tracking on Long Documents}
TDT techniques could be categorized into three types: clustering, classification, and topic modeling.
Many clustering approaches treat the detection and tracking tasks as one 
\cite{allan1998topic,yang1998study,brants2003system}: using clustering and content similarity to segment existing document collection into stories (i.e., detection) 
and mapping new articles to one of the stories (i.e., tracking).
When no similar story is found for the article, a new story is detected.
However, the word ambiguity is often overlooked in these methods and the focus is on analyzing the larger clusters.
Stories that are less popular (small clusters) are either annexed by large clusters or filled with noisy content from the popular stories.

The supervised TDT approaches generally assume a static environment \cite{makkonen2004simple}. 
A single classifier is typically trained off-line on a relatively small batch of manually labeled data. 
The classifier is then deployed for detecting events directly or combined with a clustering approach.
In other works, Kumaran et al. \cite{kumaran2005using} and Zhang et al. \cite{zhang2007new} employ the named entities detection techniques and enhance the feature vector space with named entities in the topic detection task.

Probabilistic topic modeling \cite{alsumait2008line,leek2002probabilistic} has been successfully applied to predict the underlying topics of documents.
Takahashi et al. \cite{takahashi2012applying} first use a dynamic topic model to learn the topics from a news stream, and then apply Kleinberg's model  \cite{kleinberg2003bursty} to detect the bursty topics.
The work in \cite{wang2007mining} builds a topic model to find bursty topics from text streams. %structure of text.
%The model learns the distributions of underlying
However, these techniques are restricted in scope, because limited labeled data is available for training, especially in a dynamic environment. 
For instance, new terms, new persons and new developments of the story may emerge. 
Static models for tracking stories are prone to both false positive and negative errors when a concept drift occurs in the data stream. 
Techniques such as incremental learning and ensemble methods can be employed to account for unseen events and adapt to changes that may occur over time \cite{Morinaga:2004:TDT:1014052.1016919,alsumait2008line}.

\subsection{Topic Detection and Tracking on Tweets} 
With the rising of the social media platforms, many recent works focus on TDT on Twitter data.
In these works, a collection of bursty terms that represent a topic are usually detected from the document stream and used later on for topic tracking. The bursty terms could be in the format of keywords \cite{cataldi2013personalized,he2010topic,mathioudakis2010twittermonitor,sakaki2010earthquake}, hashtags \cite{alvanaki2012see,feng2015streamcube}, phrases \cite{leskovec2009meme} or segments \cite{li2012twevent}.
Petrovic et al. \cite{petrovic2010streaming} examine the first-story detection problem in tweet streams, and propose an algorithm using locality-sensitive hashing to overcome scalability issues of traditional approaches.
Lin et al. \cite{lin2011smoothing} propose a broader definition of topics that might include a disparate collection of loosely-related events, and build language models for each topic to filter the entire tweet stream. 
They explore the simplest possible classifier based on computing the perplexity of unseen tweets.
The work in \cite{magdy2014adaptive} presents an adaptive method for following dynamic topics on Twitter. 
The system starts with topics that are expanded with a rich set of hand tuned keywords. These human-generated keywords are used by the classifier for judging relevance of the incoming tweets. 
Wang et al. \cite{wang2013exploiting} propose the use of hashtags for adaptive microblog crawling. They show that their adaptive algorithm identifies more relevant tweets as compared to a traditional pre-defined keyword classifier.
The work in \cite{yin2013unified} takes a topic modeling approach and creates a unified model to distinguish temporal topics from stable topics.

\subsection{TREC Microblog Track}
The Microblog track at TREC \cite{ounis2011overview,soboroff2012overview,lin2014overview} started with the purpose of evaluating search methodologies in microblogging environments like Twitter.
It studies the problem of topic tracking in Twitter, which aims at filtering information relevant to a given query (short phrases) from real-time tweet streams.
The winning team of TREC2014 \cite{lv2014pkuicst} applies a L2R framework with semantic score features, semantic expansion features and document quality features. 
Along with query expansion and document expansion techniques, the approach computes the relevance score of a given topic and tweet from different perspectives.

In this paper, we combine ideas from topic tracking and entity disambiguation, and propose an approach that tracks the target story across different domains using the semantic structure of the story.
We employ an entity disambiguation solution to correctly interpret the underlying meaning of the text, and represent the story in a high-level entity graph.
Unlike the traditional topic tracking techniques, we model story tracking as a L2R task by classifying story-doc pairs as relevant or irrelevant.
We compare our approach to 6 SOTA story tracking approaches and show that we outperform the F1 of the traditional topic tracking approaches in mixed sources streaming environment using fewer labeled data.
%Our experiment results show that we outperform the accuracy of the traditional topic tracking approaches using much fewer labeled data, 
%especially when tracking complex news stories where a rich amount of the story details are reported in the news.

%\section{Related Work}\label{sec:rel_story}

\section{Story Disambiguation Framework}
\label{sec:story-method}
%!TEX root = main.tex

In this section, we present the technical details of our Story Disambiguation framework. 
We start with the overview of our method and the key terminology, and then discuss the 
main algorithms in our approach.

\subsection{Method Overview}

\begin{figure}[tbh]
    \centering
    \includegraphics[width=0.9\linewidth, ]{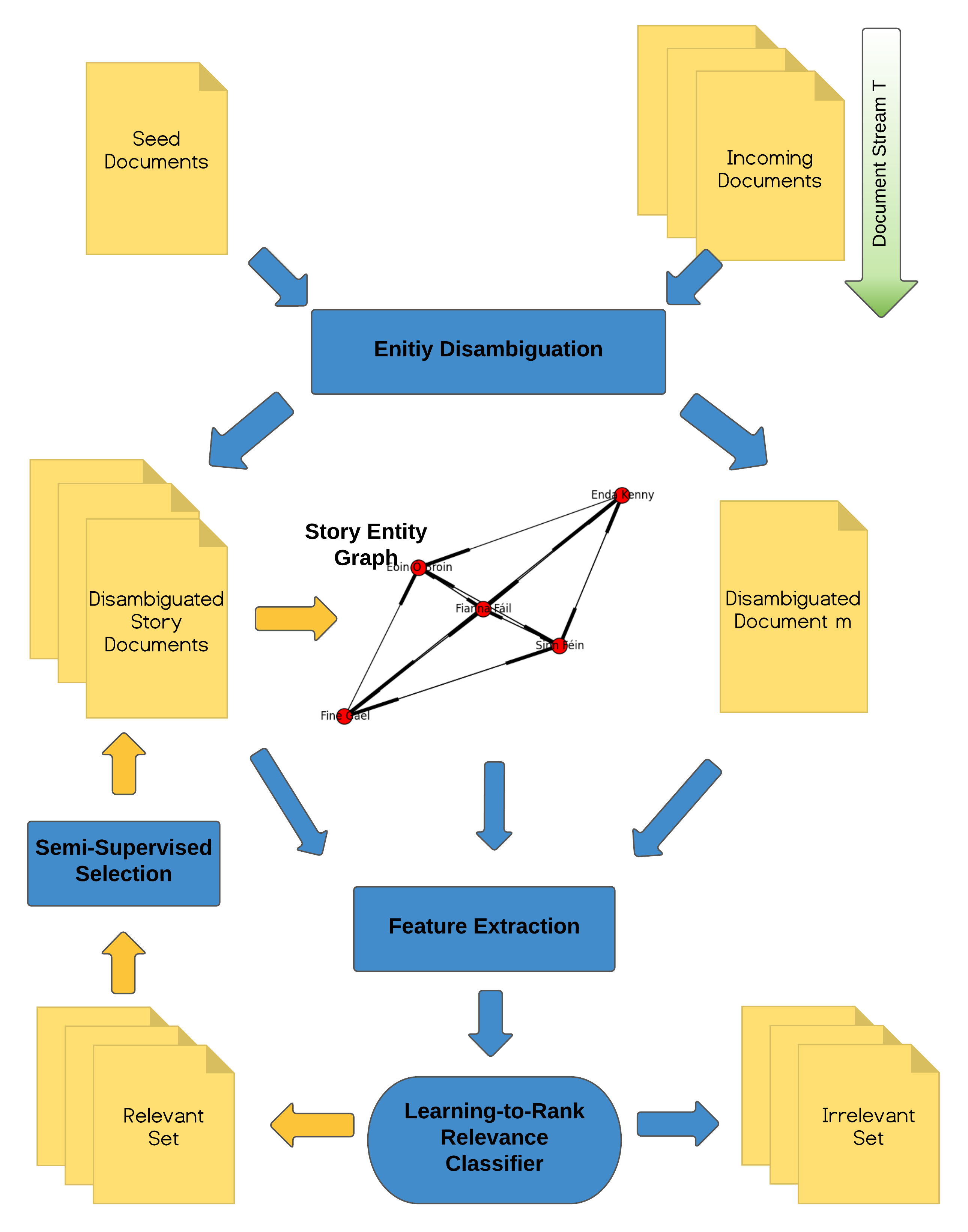}
    \caption{Story Disambiguation framework: high level overview.}
    \label{fig:high-level-overview}
%    \vspace{-0.4cm}
\end{figure}

The goal of the Story Disambiguation framework is to classify streaming documents (e.g., news articles, comments, tweets) as to whether or not they belong to a target news story.
A high-level overview of our approach is shown in Figure \ref{fig:high-level-overview}.
The top half of the figure shows the construction of the story representation, including both the \emph{story docs} and the \emph{story entity graph},
 while the bottom half presents the process of classifying new documents into relevant or irrelevant to the story. \textbf{Algorithm \ref{alg:sd}} 
 presents the detailed steps of our framework.
We first merge the document streams into one stream $T$, the content of which is sorted in chronological order (top right corner in Figure \ref{fig:high-level-overview}).
The stream $T$ mixes documents from different domains, e.g., news articles and tweets.
%The documents in $T$ are processed sequentially in chronological order.
The target story $S$ is initially defined by the user, with a few \emph{seed docs}, which could be from a single domain (i.e., articles), or from mixed domains (i.e., articles and tweets).
Our algorithm pre-processes (e.g., entity disambiguation) and iteratively expands (e.g., semi-supervised selection) the set of seed docs 
into a new set called the \emph{story docs}. For brevity, we refer to the set of documents describing the target story at any point in time 
as the  \emph{story docs}.

At tracking start time $t_0$, we first \textbf{disambiguate all story docs} and map a selection of text phrases to corresponding 
Wikipedia articles (top of Figure \ref{fig:high-level-overview}), as also detailed in Section \ref{sec:ned}.
Next, we \textbf{construct a story entity graph} $G_{t_0}$ for the target story $S_{t_0}$, with the entities from the story docs (center of Figure \ref{fig:high-level-overview}). 
The nodes of the graph are the entities, while the edges are their co-occurrence in the story docs.
More details regarding the entity graph are presented in Section \ref{sec:entity-graph}.
We train an L2R model that captures the relevance of arbitrary pairs of story and document objects.
We explain the \textbf{L2R model training and feature extraction} in Section \ref{sec:classifier-feature} - \ref{sec:feature-engineering}.

At the current time $t_c$ (where $t_c >t_0$) the L2R model classifies the relevance of an incoming document $m \in T$, to the current story $S_{t_c}$.
The features are extracted from the entity graph $G_{t_c}$, the story docs and the disambiguated document $m$.
Depending on the classification result, we add the classified document $m$ to one of two sets, the relevant set $R$ or the irrelevant set $I$ (bottom of Figure \ref{fig:high-level-overview}).
Then we update the story entity graph $G_{t_c}$ and the story docs with the documents in the relevant set $R$ using \textbf{semi-supervised selection}, 
a new method we present in Section \ref{sec:semi-supervised}.

The method continues the classify-update cycle until the end of the stream $T$.
In these cycles, the story entity graph $G$ and story docs are continuously updated, representing the updates in the news story.
The feature vectors will reflect the story drift over time, but the L2R model, which captures the relationship between features and outcome (i.e., relevance) remains static from the beginning to the end of the story tracking.
%The overview of the Story Disambiguation model is presented in Algorithm \ref{alg:sd}.
{\LinesNumberedHidden
    \begin{algorithm}[tbh]
    \begin{small}
        \caption{Story Disambiguation Model}
        \label{alg:sd}
        {\bf Input}:     
        Seed document set $Seed$, a stream of documents $T$, 
        pre-trained L2R model $c$.
        
        {\bf Output}:
        Relevant documents set $R$.
        
        {\bf Method}:
        
        $R=\emptyset$; 
        $StoryDoc=\emptyset$ {\scriptsize //Set \emph{story docs} to empty.}
        
        $Graph=\emptyset$ {\scriptsize //Set entity graph $G$ to empty.}
        
        GLOBAL $SSS_{cycle} = 0$ {\scriptsize //$SSS_{cycle}$ records number of semi-supervised selection cycles.}
        
        \For{$s\in Seed$}{
            {\scriptsize //Disambiguate seed documents}
            
            $s_{NED} = EntityDisambiguation(s)$
            
            $s_{NED}$ marked as $Unavailable$.
            
            $Seed_{NED} = Seed_{NED} \cup s_{NED}$
        
        }
        
        $StoryDoc = Seed_{NED}$ {\scriptsize //Add seed documents to \emph{story docs}}
        
        $Graph = ConstructGraph(StoryDoc)$ {\scriptsize //Construct story graph $Graph$ based on $Seed$ documents. Presented in Alg \ref{alg:GC}.}

        \For{$m\in T$}{
            {\scriptsize //Classify incoming document $m$ from stream $T$}
        
            $m_{NED} = EntityDisambiguation(m)$
            
            $m_{NED}$ marked as $Available$
            
            $f = FeatureExtraction(m_{NED},StoryDoc,Graph)$ 

            $p = L2RModel(c, f)$ {\scriptsize //Classify $m$ as relevant or irrelevant, get predicted probability $p$}
            
            \uIf{$p >= 0.5$}{
                {\scriptsize //If probability above 0.5, put $m$ in relevant set $R$}
                
                $R = R \cup m_{NED}$
            
            }
            \Else{
                $I = I \cup m_{NED}$
            }
        
            \If{ found 50 new documents in $R$}{
                {\scriptsize //Start semi-supervised selection cycle. Presented in Alg \ref{alg:AR}.}\\
%                {\scriptsize //Presented in Alg \ref{alg:AR}.}\\
                $StoryDoc,Graph,R,I = UpdateStorySSS(StoryDoc,Graph,R,I)$ 
            }
        }
        {\bf Return $R$}
       \end{small}
    \end{algorithm}
}

\subsection{Preprocessing Text with Entity Recognition and Disambiguation}\label{sec:ned}

The goal of entity recognition and disambiguation is to identify named entities (e.g. people, locations, organizations) in a given text, 
and map potentially ambiguous entities, to their canonical representation in an external knowledge base (e.g., Wikipedia entries). 
In this paper, we choose Wikipedia as the external knowledge base. After entity recognition and obtaining a list of named entities from the text, we disambiguate them by mapping them to Wikipedia article ids, along with a confidence score. 
%Only the entities with confidence score higher than a pre-defined threshold are used in our experiments.

We extend the entity recognition techniques proposed in \cite{shi2017hashtagger} as follows. Given a document, we retrieve an initial set of named entities using the NLTK Python library\footnote{\url{http://www.nltk.org}}, 
then we parse the text with a POS-tagger and retain only the proper nouns, frequent nouns and common nouns. 
We find a list of noun phrases by grouping adjacent nouns together as one entity, and append them to the named entity list. 
Each entity in the list is annotated with its location in the text, and the same entity can appear multiple times in one document.

%Unfortunately, due to commercial license limitations, we could not use the work presented in Chapter \ref{chap:entity_disambiguation}.
We use the open source entity disambiguation solution TAGME\footnote{We used TAGME1.8 Web API \url{https://sobigdata.d4science.org/web/tagme}.}, 
which has a fast query response time as well as good accuracy.
%TAGME constructs a relation graph of sense-sense and sense-mention with the edge weights representing relatedness, where the mention refers to a text phrase in the text, and sense refers to the candidate meaning of the phrase in a knowledge base.
%In the experiment in Chapter \ref{chap:entity_disambiguation}, we find 
%, making it the second best choice in our real-time, large scale problem setting (the best choice is our proposed method in Chapter \ref{chap:entity_disambiguation}.). 
We send the list of named entities to the TAGME API, and retrieve corresponding Wikipedia IDs. 
We create fake Wikipedia IDs for the entities that do not have corresponding Wikipedia entries, 
or are mapped with low confidence score (i.e., mapping probability lower than 0.5), 
so the story entity graph is not entirely dependent on the quality of the TAGME results.

When processing short content such as tweets, we found that TAGME is not always accurate.
Some story entities that are important for the tweets, but are less frequently used in general, are often wrongly disambiguated or disambiguated with low confidence scores.
To improve the disambiguation quality for short texts, we group tweets tagged with the same hashtags and posted together within a short time-window (i.e., 1 hour), and disambiguate them as one single document.
Tweets with the same hashtag are usually discussing the same topic, and thus could act as surrounding context for each other.
Since TAGME employs a collective ranking approach (i.e., all mentions in the text are disambiguated together) it can take advantage of the longer context to provide better disambiguation accuracy. 
For tweets without hashtags, or other short/medium length text (e.g., chats, comments), one possible solution is to first use hashtag recommendation techniques \cite{shi2017hashtagger}, 
then disambiguate the hashtagged content as above. %similarly to hashtagged tweets.
Such technique is not applied in this paper.
%We would like to test such approach in future experiments.

\subsection{Story Entity Graph}\label{sec:entity-graph}
This section presents the construction and update of the story entity graph $G$, and describes the weighting and ranking of story entities using the graph.

\textbf{Graph Construction.} 
The entity graph for a tracked story is a weighted, undirected graph.
For each document in \emph{story docs}, we represent the text as a list of disambiguated entities, 
with three components: the entity id (Wikipedia article id), the positions in the text (measured in characters from the start), 
and the disambiguation confidence score returned by TAGME.
For documents that have structure (e.g., titles, subtitles), we flatten their structure (i.e., the title will be the first sentence of the article).
Then we create a node for each new entity, and add the node to the entity graph $G$.

We create the edges of the graph based on entity co-occurrence in the individual documents of the story docs.
The most valuable information in news content is typically embedded at the beginning of the article.
For example, the title of a news article often summarizes the entire article. 
Therefore, the entities appearing at the beginning of the article are more important.
To emphasize their importance in the graph, we create a dynamic sliding text-window $tw$, the length of which depends on the entity's position $p$ in the document $m$.
At the beginning of the document, the sliding text window $tw$ has the longest length (as measured in characters).
The length decreases as $tw$ moves towards the end of the document.
Equation \ref{eq:tw} gives the formula for computing $tw$.
Parameters $a$ and $b$ are set empirically to 500 and 400 respectively. 
%Why these values, any intuition regrding how many words the window captures?
%Do you slide one char at a time, or by 500 chars?
Thus in the beginning of the document we create an edge for any two entities co-occuring within 500 characters ($\sim$50 words), while towards the end of the document, we 
shrink the text window to 100 characters ($\sim$10 words). 
%Thus, the dynamic text window $tw$ emphasizes proximity of entities in different parts of the text. 
%For example, if entities co-occur in the beginning of the text, 
%the window $tw$ is longer, 
%(i.e., the number of characters between two entities smaller than a dynamic threshold)
For any two entities that appear together in $tw$, 
we create an edge between the corresponding nodes in the graph $G$, with initial edge weight $1$.
If an edge already exists in the graph, we increase the edge weight by $1$.
%Note that there could be more than one connections between two entities per document, if they co-occur in multiple dynamic time windows.
%From our previous study in Chapter \ref{chap:hashtagger}, 
\begin{equation}
\label{eq:tw}
|tw| = a - b \times \frac{p}{|m|}
\end{equation}
%where $a$ and $b$ are two variables.

Because we create graph edges based on the entity co-occurrence in the dynamic text-window $tw$, 
we ensure that entities appearing at the beginning of the article, have more edges with higher edge weights, 
and thus, reflect their central role in the news stories.

\textbf{Graph Update.} Updating the story graph when adding new documents to the story docs, is done by adding nodes and edges to the graph, 
in a similar process to the graph construction.
When removing documents from the story docs, we reduce the affected edge weights by 1, then remove edges with weight 0 and nodes with no edges.
This very efficient way of updating the entity graph is essential for our problem setting, 
because we update the entity graph frequently, and the size of the graph can be in the order of few thousand nodes and tens of thousand edges.

\textbf{Biased PageRank for Weighting and Ranking Entities.}
The weights of nodes in the story entity graph $G$ represent the importance of the corresponding entities in the target story $S$. 
To appropriately weight the nodes, we employ a biased PageRank algorithm.
PageRank \cite{page1999pagerank} is a link analysis algorithm originally designed to measure the importance of web pages.
It works by counting the number and quality of incoming links to a page, with the assumption that more important web pages are likely to receive more links from other pages.
When edges are weighted, PageRank considers higher weights are better for incoming edges.
%PageRank can also work on a undirected graph. 
%One way is to consider each undirected edge as two directed edges with same weights.
Different biased PageRank algorithms are introduced when extra information is used in ranking. 
For example, personalized PageRank \cite{chakrabarti2007dynamic} biases to a specific set of pages according to a user's preferences.
Instead of performing a random jump, personalized PageRank jumps only to a set of user defined nodes.
On the other hand, topic-sensitive PageRank \cite{haveliwala2002topic} biases to particular topics of interest and pages of relevant topics have higher probabilities when jumping.
TrustRank \cite{gyongyi2004combating} biases to a set of trustworthy domains to fight spam. 
%The pages from trustworthy domains have higher initialization weights in TrustRank.

We use biased PageRank to compute the weights of the nodes in the story entity graph $G$ based on the structure of their edges. 
In our problem setting, our bias is the original story \emph{seed docs} provided by the user to define the story.
Regardless of whether the story drifts, the focus of the story entity graph should be close to the original \emph{seed docs}.
Thus, we employ the same method as the personalized PageRank: 
We jump back to one of the top 10 nodes with highest node weights in the first story entity graph $G_{t_0}$, which is constructed from the \emph{seed docs}.
After computing weights for nodes, we rank the nodes and corresponding entities by their weights.
Algorithm \ref{alg:GC} presents the process of entity graph construction and weights computation.
We use the NetworkX Python library\footnote{\url{https://networkx.github.io/}} to construct the story entity graph and to compute 
biased PageRank \cite{langville2005survey} weights.

{\LinesNumberedHidden
    \begin{algorithm}[tbh]
    \begin{small}
        \caption{Graph Construction and Entity Ranking via Personalised PageRank.}
        \label{alg:GC}
        {\bf Input}:     
        Set of disambiguated story documents $StoryDoc$; Set of disambiguated seed documents $Seed_{NED}$.
        
        {\bf Output}:
        Story entity $Graph$ with entity weights.
        
        {\bf Method}:
        
        $Graph=\emptyset$ {\scriptsize //Entity graph is empty.}
        
        \For{$m_{NED}\in StoryDoc$}{
        
            {\scriptsize //For each disambiguated document in $StoryDoc$.}
            
            \For{$e \in m_{NED}$}{
                {\scriptsize //For each entity $e$ in document $m$}

                $tw = a - b \times \frac{e_{pos}}{|m|}$ {\scriptsize //The length of text window depends on the position of $e$}
                
                \For{$e\prime \in m_{NED}$ and $e\prime_{pos} > e_{pos}$ and $e\prime_{pos} - e_{pos} <= tw$}{
                    {\scriptsize //For entities $e\prime$ occuring after $e$ in document $m$, within $tw$, create nodes and edges in graph.}
                        
                        $Graph = CreateNode(e)$
                        
                        $Graph = CreateNode(e\prime)$
                        
                        $Graph = CreateEdge(e,e\prime)$
                
                }    
            
            }
            
        }
        
        %$Seed_e = SelectKeyEntities(Seed_{NED})$ {\scriptsize //Select top 10 nodes from $Seed$.}
        {\scriptsize //Compute entity weight using PageRank.}\\
        $Graph = PersonalizedPageRank(Graph,Seed_{NED})$ 
        
        {\bf Return $Graph$}
       \end{small}
    \end{algorithm}
}

%\todo{Add a figure to show an example of text to graph then to node weights process}
\subsection{Modeling Story Relevance with Learning-to-Rank}\label{sec:classifier-feature}

%In this section, we present the feature engineering and the training for the relevance L2R model.
%In this section, we present our learning model for Story Disambiguation.
%Similar to the setting presented in Chapter \ref{chap:hashtagger}, w
We employ a Learning-to-Rank (L2R) approach to model story relevance. 
%We first remind the reader of the fundamental ideas of L2R and then present our proposed model.
In an Information Retrieval L2R setting, given a collection of documents $C$, and a query $q$, a global L2R model retrieves a list of documents $C\prime$ ranked by their relevance to the query $q$.

In the story tracking setting, we consider stories as queries. Given a story, we want to retrieve and rank documents that are relevant to the story.
The difference between our problem setting and the traditional L2R problem is that: 
1) our document collection is a stream, rather than a static set; 
2) our query is dynamic (i.e., story drift) and has structure (i.e., much more complex than keyword queries); 
3) for story tracking, we care more about finding all relevant documents and less about fine grained levels of relevance (i.e., we focus on the relevant/irrelevant split).

We use pointwise L2R for our setting, since it is efficient in a streaming environment 
and works well for the binary relevance setting \cite{Agarwal2012,shi2017hashtagger}.
We train a pointwise L2R model to classify arbitrary story-doc pairs as to whether these pairs are relevant or irrelevant.
The story-doc pairs are represented by a feature vector, where the features are extracted from the story representation 
(i.e., entity graph and story docs) and the document representation (i.e., the document content and the disambiguated entities). 
We present the feature engineering process in Section \ref{sec:feature-engineering}.

\textbf{Model Training.} 
After comparing several well-known classifiers, we select RandomForest as our binary relevance classifier.
%(these experiments are presented in Section \ref{sec:compare-classifiers}).
To train the classifier, we collect news articles and tweets for an example news story, 
namely the ``Irish General Election 2016" (GE16), and use them as training data.
We collect 8 other news stories as testing data.
Details regarding the data collection are presented in Section \ref{sec:label-data}.
In an ideal situation, we should use multiple stories and their relevant text documents as training data, 
to generalize the relationships between different types of documents and news stories (e.g., stories with various life span, popularity, noise level, complexity, and locality).
However, since we did not find an existing dataset of stories with multiple types of documents, and manual labeling is very costly, we only use one story, GE16, for training and validating parameters.
We test the classifier on the other 8 news stories to evaluate its performance.
%Why no train and test on different story types?
To simulate the effects of having training story-doc pairs gathered from different stories, we generate multiple ``stories'' using only the GE16 story data as follows.
We randomly select some documents (a mix of articles and tweets) from the GE16 data, and consider them as the \emph{story docs}.
A story entity graph $G$ is created based on the \emph{story docs}.
Then, we randomly select 0.9k articles and 1.8k tweets about GE16, and also 10 times more irrelevant documents (27k) from other news stories in the same period, to simulate the text stream $T$.
Thus, we get a story representation and multiple incoming text representations with relevant/irrelevant labels. 
We pair the story to all documents in $T$, and extract features from these pairs.
We repeat the above process 15 times, generating 15 distinct story representations with different sizes for \emph{story docs} and the entity graph $G$.
In total, we get 445.5k ($29.7k \times 15$) story-doc pairs, and we train the RandomForest classifier with these training pairs.
Further details on the training data generation are presented in Section \ref{sec:training}.

\subsection{Feature Engineering}\label{sec:feature-engineering}

\begin{table*}[]
\centering
\caption{The 14 features for the story relevance L2R model, extracted from the story entity graph $G$, 
\emph{story docs} set, and the document $m$. ``\#'' refers to the ``number of'' in this table. The title of $m$ refers to the first 140 characters in $m$, while the body of $m$ refers to the content after that.}
\label{table:features}
%\resizebox{\textwidth}{!}{%
\begin{tabular}{|c|c|l|l|}
\hline
Category                                     			& Type                         	& Feature                       & Description                                                                             \\ \hline
\multicolumn{1}{|c|}{\multirow{2}{*}{Story}} 	& \multirow{2}{*}{Text} 	& 1. $|$\emph{story docs}$|$                  & \# documents in \emph{story docs}                                                    \\ \cline{3-4} 
\multicolumn{1}{|c|}{}                       			&   					& 2. $|G|$                           & \# nodes in $G                                                       $ \\ \hline
\multicolumn{1}{|c|}{\multirow{2}{*}{Document}} & \multirow{2}{*}{Text}  	& 3. $|m|$                           & \# characters in $m$                                            \\ \cline{3-4} 
\multicolumn{1}{|c|}{}                       			&                              		& 4. $|m_{e}|$                     & \# entities in $m$                                                          \\ \hline
\multirow{10}{2cm}{Story \& Document}                & \multirow{6}{*}{Text}  	& 5. $|Overlap_{t}|$    & \# entities in both  \emph{story docs} and title of $m$ \\ \cline{3-4} 
                                             				&                              & 6. $|Overlap_{b}|$     & \# entities in both \emph{story docs} and body of $m$                         \\ \cline{3-4} 
                                             				&                              & 7. $|Overlap|$          & $|Overlap_{t}| + |Overlap_{b}|$                              \\ \cline{3-4} 
                                             				&                              & 8. Avg($|Overlap|$) & $|Overlap|$ divided by the $|m_e|$                                                 \\ \cline{3-4} 
                                             				&                              & 9. CosSimi\_True          & Cosine similarity between \emph{story docs} and $m$                                       \\ \cline{3-4} 
                                            
                                             				&                              & 10. CosSimi\_False         & Cosine similarity between irrelevant sample from stream and $m$                                       \\ \cline{2-4}
                                             				& \multirow{4}{*}{Graph} & 11. $Simi_{t}$       & Sum of the node weights of $Overlap_{t}$       \\ \cline{3-4} 
                                           				&                              & 12. $Simi_{b}$        & Sum of the node weights of $Overlap_{b}$           \\ \cline{3-4} 
                                           				&                              & 13. $Simi$            & $Simi_{t} + Simi_{b}$ \\ \cline{3-4} 
                                             				&                              & 14. Avg($Simi$)     & $Simi$ divided by $|m_e|                                                       $ \\ \hline
\end{tabular}%}
\end{table*}
In general, the features for a L2R model can be categorized into 3 groups: the features that represent the query (i.e., story) alone, 
the document alone, and those that represent the association between the query and the document (i.e., story and document).
For a relevance model, the features that capture the relationship between the query and the document are the most important,
 while the features from the other two groups also help to characterize the document-query pairs, e.g., by considering the document length, the query popularity, etc.

Depending on how the features are extracted, we can also split each feature category into graph-based features and text-based features.
The graph-based features are extracted from the story entity graph $G$, while the text-based features are from the \emph{story docs}, incoming stream document $m$, and a set of irrelevant documents sampled from the stream.
We further split document $m$ into 2 parts: document title $m_t$ and document body $m_b$, by a fixed 140 character threshold (set based on tweet length constraint).
%This number is chosen according to the maximum length of tweets.
The graph-based features include the graph size and the entity weights within the graph.
The entity weight captures the importance of that entity in the story, a smaller weight means a lower importance rank.
The text based features include the text length, the cosine similarity between the tf-idf profiles of story and target documents, and also involve the disambiguated entities in the text.

Table \ref{table:features} presents the features for the L2R model in detail.
Among the 14 features, 4 of them capture the characteristics of the story or the document alone.
For the other 9 features, one is the cosine similarity between the tf-idf profiles of the \emph{story docs} and the target document $m$, another one is the cosine similarity between the irrelevant seed sampled from the stream and $m$.
The other 4 $Overlap$ features measure the entity overlap between the disambiguated \emph{story docs} and the disambiguated target document $m$.
The last 4 $Simi$ features measure the similarity between the entity graph $G$ and the document $m$ by weighting the $Overlap$ features using the node weights of the story graph $G$.

We also tried more sophisticated graph similarity features, by turning the document $m$ into a small graph and comparing it to the story graph $G$.
However, through experiments, we found that such graph similarity features do not work well, when compared to using the entity weights of $G$ directly.
This is because the graph similarity measures the edge overlap, instead of node overlap between two graphs.
For a target document to be relevant to a news story, it is important that the document mentions the key entities of the story (e.g., people, locations), while whether two entities co-occur in a text-window is less relevant.
On the other hand, the entity co-occurrence (i.e., edges) are important when weighting and ranking the nodes (i.e., entities) in the entity graph $G$, and thus, support the entity similarity features $Simi$. 

\subsection{Semi-Supervised Selection}\label{sec:semi-supervised}

Since the user only provides a limited number of \emph{seed docs} to define a story initially, one of the challenges for the proposed Story Disambiguation framework is to enrich the \emph{story docs} and the story entity graph $G$ by adding new information as the story develops over time. 
%so we could correctly reflect the PageRank development at each time point. 
We use Semi-Supervised Learning (SSL) to select some incoming documents, which are weakly labeled by our model, for expanding the story representation. 
{\LinesNumberedHidden
    \begin{algorithm}[t!]
    \begin{small}
        \caption{Semi-Supervised Selection: Accumulate+Revisit.}
        \label{alg:AR}
        {\bf Input}:     
        Story document set $StoryDoc$, entity graph $Graph$, 
        relevant set $R$, Irrelevant set $I$.
        
        {\bf Output}:
        Updated $StoryDoc$, $Graph$, $R$, $I$.
        
        {\bf Method}:
        
        \If{$SSS_{cycle} \mod 10 \ne 0$}{
        {\scriptsize //Accumulate cycle}
        
            \For{$m_{NED} \in R_{new}$}{
                {\scriptsize //For latest 50 documents in $R$}
                
                \If{$p_{m} >= 0.8$}{
                    {\scriptsize //Add document $m$ with predicted probability above 0.8 to $StoryDoc$ and update $Graph$}
                    
                    $StoryDoc = StoryDoc \cup m_{NED}$
                    
                    $Graph = UpdateGraph(Graph,StoryDoc,Seed_{NED})$
                
                }
            
            }
            
            $SSS_{cycle} += 1$           
        }
        \If{$SSS_{cycle} \mod 10 = 0$}{
            {\scriptsize //Revisit cycle.}
            
            \For{$m_{NED} \in R\cup I$  and $m_{NED}$ is $Available$}{
                {\scriptsize //Re-classify document $m$ based on current $StoryDoc$ and $Graph$}
                
                $f = FeatureExtraction(m_{NED}, StoryDoc, Graph)$
            
                $p = L2RModel(c, f)$

                \uIf{$p >= 0.5$}{
                    {\scriptsize //Put document $m$ in $R$, remove it from $I$}
                    $R = R \cup m_{NED}$
                    
                    $I = RemoveFromSet(I,m_{NED})$

                }\Else{
                    {\scriptsize //Put document $m$ in $I$, remove it from $R$}
                    $I = I \cup m_{NED}$
                    
                    $R = RemoveFromSet(R,m_{NED})$
             
                }
                
                \uIf{$p of m_{NED} is the highest and p >= 0.8$}{
                    {\scriptsize //Put document $m$ in $StoryDoc$}
                    
                    $StoryDoc = StoryDoc \cup m_{NED}$
                    
                }\Else{
                    {\scriptsize //Remove document $m$ from $StoryDoc$}    
                    $StoryDoc = RemoveFromSet(StoryDoc,m_{NED})$
                
                }
            
                $m_{NED}$ marked as $Unavailable$.
                
                $Graph = UpdateGraph(Graph, StoryDoc,Seed_{NED})$
                
            }
            
            $SSS_{cycle} += 1$
            
        }
                
        {\bf Return $StoryDoc, Graph,R,I$}
    %\end{algorithmic}
       \end{small}
    \end{algorithm}
 
    %\vspace{-0.5cm}
}
\textbf{Unlabeled Data Selection in SSL.}
%% SSL definition & references
Semi-Supervised Learning (SSL) \cite{ChapelleSchZien06,Seeger02,ZhuGoldberg09} does not seek to eliminate the need for supervised labels completely, but resorts to unlabeled data to minimize the amount of labeling effort required to achieve good performance. In SSL approaches, a large amount of unlabeled data $U$, together with labeled data $L$, is used to build better classifiers.
However, it is also well-known that using all data in $U$ is not always helpful for SSL algorithms. Therefore, to select a small amount of useful unlabeled data  $U_S$, various sampling (and selection) techniques have been proposed in the literature, including the self-training \cite{McCloskyCharniakJohnson08,RosenbergHebertSchneiderman05,Yarowsky95} and co-training \cite{BlumMitchell98} approaches, confidence-based approaches \cite{LeKim142,LeKim14,MalKumJinJainYi14}, 
density/distance-based approaches \cite{DagliRajaramHuang06,ReitmaierSick11},
and other approaches used in active learning (AL) algorithms
\cite{DonmezCarbonellBennett07,FuLiZhuZhang14,HeJiBao09,LengXuQi13,Settles10}.

The goal of unlabeled data selection is to iteratively improve the performance of a supervised learning model, 
%especially for boosting algorithms, 
by using highest confidence (e.g., predicted probability) samples $U_S$ from $U =
\{(x_i)\}_{i=1}^{n_u}$ to extend the labeled data $L = \{(x_i,y_i)\}_{i=1}^{n_l}$, where $x_i \in \mathbb{R}^{d}$ and $y_i \in \{+1, -1\}$. 
At each iteration, %from boosting, 
the most confident instances are selected from $U$ and then provided for the retraining of the supervised classifier. 
After repeating the select-and-train process, the performance of the classifier improves due to the extra training data from $U_S$.

\textbf{Unlabeled Data Selection in Story Disambiguation.}
We learn from prior SSL approaches as well as unlabeled data selection techniques, and propose new selection approaches to enrich the story representation in our Story Disambiguation framework.
Similar to SSL, we have very few labeled data $L$ (i.e., \emph{seed docs}) to define a story, and a huge number of unlabeled data $U$ from the document stream $T$.
Unlike the static datasets in most SSL problem settings, our unlabeled data $U$ is a stream, where the data is gradually available to us, and the recent data in $U$ is more valuable than the old.
In addition, we do not retrain our classifier after expanding the \emph{story docs} set with unlabeled documents.
We only update the feature vectors to reflect the changes in the \emph{story docs} and the corresponding entity graph $G$.

To select the highest confidence samples $U_S$ from $U$ iteratively, there are two styles of approaches: 1) \textbf{\emph{Accumulate}}  and 2) \textbf{\emph{Revisit}}.
\emph{Accumulate} means the selected data $U_S$ in each iteration is accumulated.
Once an unlabeled example is selected in one iteration, it will be used to retrain the classifier in each following iteration.
The \emph{Revisit} style does the opposite, at every iteration, the previous $U_S$ is discarded and re-selected, based on the current classifier (feature vector in our case) and the new prediction probabilities.
These two selection styles have their pros and cons.
The \emph{Accumulate} approach often introduces noise into $U_S$, especially at early iteration cycles when the classifier still suffers from insufficient training data.
The \emph{Revisit} approach avoids such problems since the old selections are discarded. However, it is slow when compared to the \emph{Accumulate} selection approach, because it revisits all data in $U$ in every iteration.
In our streaming setting, where $U$ becomes bigger and bigger, a naive \emph{Revisit} approach is not feasible.

To address the problems with \emph{Accumulate} and \emph{Revisit} selection approaches and to adjust them to our streaming setting, we propose two variant approaches: \textbf{\emph{Revisit-Recent}} and \textbf{\emph{Accumulate+Revisit}}.

\emph{Revisit-Recent} is similar to \emph{Revisit}, with the difference that it only re-selects $U_S$ from the most recent $k$ unlabeled documents in stream $T$.
It is designed to adapt to a streaming environment and reduce the runtime.
However, we find \emph{Revisit-Recent} is not as accurate as we expected. 
This is due to the fact that we ignore older documents that are very valuable in defining the story at all times.
Since the document stream $T$ is extremely dynamic, and unpredictable, only considering recent data is prone to errors.

\emph{Accumulate+Revisit} is the best selection approach among the four, as we show through our experiments in Section \ref{sec:ssl-compare}. It combines the advantages of \emph{Accumulate} and \emph{Revisit} approaches and achieves the best performance-runtime trade-off.
The \emph{Accumulate+Revisit} approach accumulates the selected unlabeled data $U_S$ in each iteration, but every few iterations, it revisits all available unlabeled data and re-selects $U_S$.
The re-selected $U_S$ in the revisit cycle are instead accumulated and used in every following cycle. 
All unlabeled data becomes unavailable after this step, and in the next iteration, the approach goes back to accumulation style with new data from the stream $T$. 
%In this thesis, we set $x$ to 10.
A detailed explanation of the \emph{Accumulate+Revisit} approach is given below and also summarized in Algorithm \ref{alg:AR}:

\begin{enumerate}
    \item Given some \emph{seed docs} that define a news story $S_{t_0}$, at tracking start time $t_0$, we construct a story entity graph $G_{t_0}$, and append \emph{seed docs} to \emph{story docs}.
    
    \item For each document $m$ from the stream $T$, we construct feature vectors using the current graph $G$, \emph{story docs} and $m$. The L2R model then classifies the documents $m$ into relevant set $R$ and irrelevant set $I$, and provides predicted probabilities. Documents $m$ are marked as \emph{available}.
    
    \item Accumulate cycles: Once $50$ new documents are assigned to set $R$, 
    we add 1 new document $m$ with the highest predicted probabilities (higher than $0.8$) to the \emph{story docs} and update the story entity graph $G$, therefore 
    updating the story representation. We then wait for $50$ new documents for the next accumulate cycle.
    
    \item Revisit cycles: Every 10 accumulate cycles are followed by one revisit cycle. Based on the current graph $G$ and the \emph{story docs}, 
    we re-construct the feature vectors for all \emph{available} documents $m$, and re-classify them with new predicted probabilities. 
    We reset the \emph{story docs} to be the same as in the previous revisit cycle and append 1 document $m$ with highest predicted probabilities (higher than 0.8) 
    to the \emph{story docs}. We re-construct the graph $G$ based on the updated \emph{story docs}. 
    For all documents $m$ processed in this revisit cycle, we mark them as \emph{unavailable}. After the revisit cycle, we go back to the accumulate cycle.
    All parameters are chosen empirically.
    \end{enumerate}

By using accumulate cycles and revisit cycles in turn, we combine the advantages of both approaches.
The noisy examples added in the accumulate cycles are no longer carried in the $U_S$ forever.
They are cleared out in the revisit cycles.
The unlabeled set $U$ in the revisit cycles does not grow cycle by cycle, but the size remains in a small range.
The older examples are not ignored, instead the best ones are selected and used in all following cycles.
Our experiments show the \emph{Accumulate+Revisit} is as accurate as \emph{Revisit} but many times faster.

%\todo{Add a figure to show the process of Accumulate + revisit method}

\section{Evaluation}
 \label{sec:story-evaluation}
%!TEX root = main.tex

\begin{table*}[t]
\centering
\caption{The characteristics of the 9 collected and labeled news stories.}
\label{table:9-stories}
%\resizebox{\textwidth}{!}{%
\begin{tabular}{|l|c|c|c|c|}\hline
News Story                  & Duration (weeks) & Popularity & Locality & Complexity \\\hline\hline
Cyclone Debbie              & 1                & Low        & Global   & Low        \\\hline
Orly Airport Attack         & 1                & Low        & Local    & Low        \\\hline
Kim-Jong-Nam Death          & 8                & Low        & Local    & Medium        \\\hline
South Africa President Zuma & 52               & Low        & Local    & Medium     \\\hline
Irish Water Charges         & 52               & High       & Local    & Medium     \\\hline
EU Migrant Crisis           & 52               & High       & Global   & Medium     \\\hline
Trump Healthcare            & 21               & Low        & Global   & High       \\\hline
London Attack               & 1                & High       & Global   & High       \\\hline
Irish General Election 2016 & 4                & High       & Local    & High       \\\hline
\end{tabular}%}
\end{table*}

\begin{figure*}[]
	\centering 
    \includegraphics[width=0.9\linewidth]{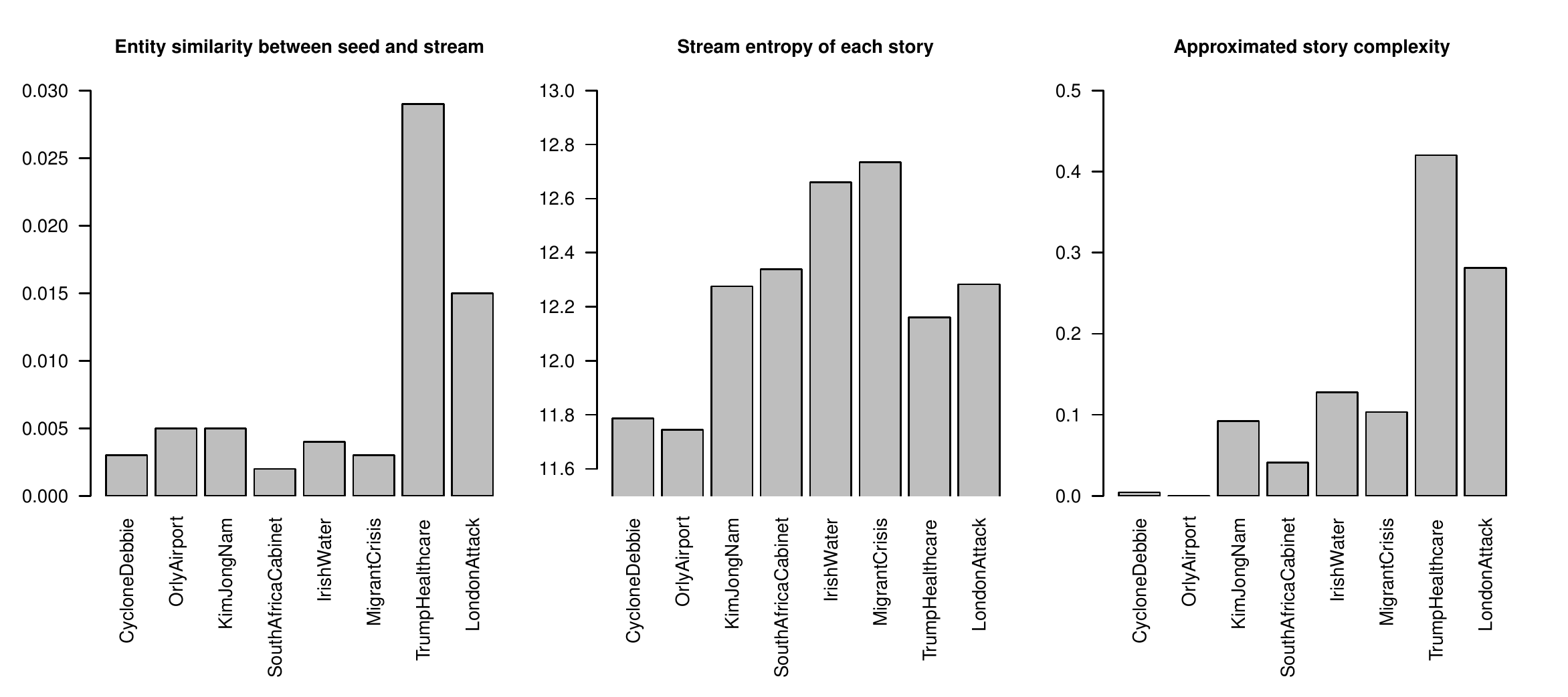}
%    \vspace{-0.4cm}
    \caption{The cosine similarity between tf-idf entity vectors for the story $seed \; docs$ and the stream $T$ (left),  the information entropy of the stream $T$ (middle), and the 
    product of the two normalized measures to obtain a single story-complexity measure (right).}
    \label{fig:story-complexity}
\end{figure*}

In this section, we discuss our methodology for gathering labeled data, and show extensive experiments analyzing the building blocks of our techniques. 
We also compare our Story Disambiguation framework to a variety of state-of-the-art (SOTA) story tracking approaches.

\subsection{Labeled Data Collection}\label{sec:label-data}
There are several labeled datasets specifically created for training and evaluating Topic Detection and Tracking (TDT) methods, 
such as TDT1\footnote{\url{https://catalog.ldc.upenn.edu/LDC98T25}}, TDT4\footnote{\url{https://catalog.ldc.upenn.edu/LDC2005T16}}, and TDT5\footnote{\url{https://catalog.ldc.upenn.edu/LDC2006T19}}. 
They consist of either news articles or the transcripts of the news broadcast, and their content is long and formal. 
In our problem setting, we would like to track diverse types of data for news stories, including formal articles and blogs, as well as noisy tweets and comments. 
Unfortunately, we did not find a dataset that contains both formal articles and social posts regarding the same set of news stories.
The existing datasets contain either one or the other data type, and combining multiple datasets is not feasible because they are not labeled with the same set of news stories.
Therefore, we create a labeled dataset using human annotators.
We first collect data from the news domain from multiple sources for over one year, and then we select 9 news stories for labeling.
We make the dataset that contains labeled news articles and tweets for each of the 9 stories available to the research community\footnote{\url{https://gitlab.com/claireshi/story-disambiguation-data}}.

\textbf{Data Collection Sources.}
As we are limited by the accessibility and labeling difficulty, we collect two distinct types of data for news stories: news articles and tweets. 
We monitor the RSS feeds of some major news publishers, including BBC, Reuters, The Irish Times, etc., for one year, and collect around 180K news articles on a variety of topics.
For the tweets, we collect data using the Twitter Streaming API with dynamically updated keyphrases that are extracted from recent news articles and reflect the current news topics.
In average, we collect around 1 million tweets per day for over a year.
For more details on this process, please see \cite{shi2017hashtagger,shi:www16}.
%The methodology for collecting news articles and tweets is similar to that used in Hashtagger+ in Chapter \ref{chap:hashtagger}.
%, and the method for extracting dynamic keyphrases is detailed in Section \ref{sec:keyphrases}.

\begin{table*}[]
\centering
\caption{The manually selected keyphrases and hashtags used to collect articles and tweets for the 9 selected stories.}
\label{table:9-stories-keyphrases-hashtags}
%\resizebox{\textwidth}{!}{%
\begin{tabular}{|P{5cm}|P{5cm}|P{3cm}|c|c|}\hline
News Story                  & Keywords/Keyphrases & Hashtags & Articles & Tweets \\\hline\hline
Cyclone Debbie              & cyclone debbie, flood, australia               & \#cyclonedebbie  & 21 & 1979   \\\hline
Orly Airport Attack         & orly, airport, paris         & \#orly & 18 & 2605 \\\hline
Kim-Jong-Nam Death          & kim jong-nam, north korea, malaysia                & \#kimjongnam    & 71 & 3518    \\\hline
South Africa President Zuma & south africa, zuma, cabinet               & \#cabinetreshufle, \#zuma     & 74 & 1005   \\\hline
Irish Water Charges         & irish water, water charges              & \#watercharges, \#right2water, \#irishwater   & 389 & 5000  \\\hline
EU Migrant Crisis           & migrant crisis, europe, eu               & \#migrantcrisis   & 809 & 5000  \\\hline
Trump Healthcare            & trump, healthcare, health              & \#trumpcare   & 42 & 3677   \\\hline
London Attack               & westminster attack, london attack               & \#prayforlondon, \#westminster, \#londonattack   & 140 & 2244    \\\hline
Irish General Election 2016 & general election, ireland, irish                & \#ge16    & 908 & 40K  \\\hline
\end{tabular}%}
\end{table*}

\begin{table*}[t]
\centering
\caption{Three settings for constructing \emph{story docs} and entity graphs $G$ for training}
\label{table:construct-graph}
%\resizebox{\textwidth}{!}{%
\begin{tabular}{|l|c|c|c|c|}
\hline
        & Articles in \emph{story docs} & Tweets in \emph{story docs} & Nodes in $G$ & Edges in $G$ \\ \hline

GE16 story 1 & 1       & 1      & 10     &  13  \\ \hline
GE16 story 2 & 1       & 2      & 29     &  84     \\ \hline
GE16 story 3 & 2       & 4      & 34     &  72   \\ \hline
GE16 story 4 & 2       & 6      & 26     &  60  \\ \hline
GE16 story 5 & 3       & 2      & 33     &  77     \\ \hline
GE16 story 6 & 3       & 5      & 48     &  121   \\ \hline
GE16 story 7 & 4       & 4      & 38     &  120  \\ \hline
GE16 story 8 & 5        & 20     & 48     &  74     \\ \hline
GE16 story 9 & 10       & 40     & 93     &  228     \\ \hline
GE16 story 10 & 20       & 80     & 205    &  658   \\ \hline
GE16 story 11 & 30       & 25     & 267    &  960  \\ \hline
GE16 story 12 & 40       & 100    & 336    &  1492     \\ \hline
GE16 story 13 & 50       & 150    & 428    &  1663   \\ \hline
GE16 story 14 & 50       & 200    & 378    &  1259   \\ \hline
GE16 story 15 & 100      & 400    & 657    &  2742  \\ \hline
\end{tabular}%}
\end{table*}

\textbf{Selecting News Stories for Labeling.}
To properly test the Story Disambiguation framework on many types of news stories, we have selected a diversity of stories with different combinations of 4 characteristics for labeling.
The attributes of news stories include:
\begin{enumerate}
    \item The \textbf{life span} of the story, varied between few days to few years.
    \item The \textbf{popularity} of the story, reflected by the number of articles and tweets related to it.
    \item The \textbf{locality} of the story: whether it attracts attention locally or globally. This is decided manually by analyzing the coverage and reach of a story. Global stories are usually popular as well.
    \item The \textbf{complexity} of the story, in terms of the entity tf-idf similarity between the story \emph{seed docs} to the stream $T$, and the $Entropy$ of the stream, as shown in Figure \ref{fig:story-complexity}. The entity similarity reflects the ambiguity of the story content, while the entropy of the stream represents how much information is embedded in the steam (higher entropy, more varied vocabulary, noisier stream content). Complex news stories tend to be more ambiguous (i.e., higher entity similarity), and the information in the stream is less focused (i.e., higher entropy).
    %Big news stories involve many people, companies, organizations, etc., while other stories are only reported with limited details and include fewer entities.
\end{enumerate}

Table \ref{table:9-stories} shows 9 selected stories with their characteristics, ordered by the complexity of the stories. 
Among the 9 stories, 3 of them only lasted for at most a week, while 3 others lasted for a year or more. 
Some stories are viral, with hundreds of articles and tens of thousands of tweets, labeled in a manner we explain in the next section. 
Half of the stories attract attention worldwide, while the other half are only of interest locally. 
Regarding the story complexity, we collected stories at three complexity levels: low, medium, and high.
Low complexity stories have low entity similarity and low stream entropy.
They only have 1 or 2 important entities (e.g., peoples' names or short phrases that appear in many of the articles and tweets of that story).
Medium complexity stories have either higher entity similarity or higher stream entropy, making the story tracking task more challenging.
The entity similarity and stream entropy are both high for complex stories, showing that the stories themselves are ambiguous, and the information in the story streams is not focused.
We show in Section \ref{sec:compare-stories} how our proposed Story Disambiguation and other SOTA methods perform on different types of stories.

\textbf{Data Collection and Labeling.} We collect relevant news articles and tweets for the 9 selected news stories.
For each story, we manually select few keyphrases/words from the story and use them in an article pre-filtering step. 
We retrieve articles with the matching keywords in their headline or subheadline from our collection, and let a human annotator label this subset of articles manually.
To ensure the tested story tracking approaches are not influenced by the keywords (which act as labels), after retrieval we \textbf{remove} the keywords from the article headline and subheadline.
The keywords filtering step significantly reduces the amount of labeling work on the annotator, reducing from the entire 180k collection down to few hundred articles per story, but with the compromise of not retrieving all relevant articles.
Regarding labeling tweets, we only label tweets as relevant to the target story if they contain the story specific hashtags (e.g., \#ge16 for Irish General Election 2016).
We identified 1 - 3 story specific hashtags, and use them as the label of the tweets.
Similarly, after retrieval, we \textbf{remove} the hashtags from the tweet.
A lot of relevant tweets that do not contain the hashtag are missing from our positive labeled set, but it is not feasible for annotators to label tens of thousands of tweets for each story.
Table \ref{table:9-stories-keyphrases-hashtags} shows the keyphrases and the hashtags we choose for the 9 selected stories and the number of relevant articles and tweets retrieved per story.
After gathering a collection of relevant articles and tweets per story as positive examples, we randomly select 10 times (1:10 positive negative:ratio) 
irrelevant articles and tweets that are published in the same time window as the story, and label them as negative examples.
Thus, we get 9 sets of labeled data, one set for each story, with only two types of labels in each set: relevant or irrelevant to that particular story.

In the following experiments, we use the Irish General Election 2016 (GE16) story as a labeled set for training and validation, and use the 8 remaining labeled story datasets as the test data.
We show the detailed experiment setup in the following sections.

\subsection{Training the Learning-to-Rank Model}\label{sec:training}

%In this section, we show the training the Story Disambiguation model on the GE16 dataset.

As discussed in Section \ref{sec:classifier-feature}, to construct story-doc pairs for training our L2R model, we prepare the story representation $S$ and the incoming text $m$ from stream $T$ separately, then pair them together.
We first create 15 story representations by randomly sampling 15 small sets of relevant articles + tweets for the \emph{story docs}, and construct the entity graph $G$ following the steps presented in Section \ref{sec:entity-graph}.
These 15 groups consist of a different number of relevant data, as shown in Table \ref{table:construct-graph}, simulating different stories.
The total number of nodes and edges in the constructed graphs vary accordingly.

We then create the document set by selecting all relevant articles ($0.9k$k) plus a random sample of relevant tweets 
($1.8$k) of the story as the positive data ($2.7$k in total), and randomly selecting 10 times irrelevant articles and tweets as the negative data ($27$k).
Thus, we have a labeled document set of size $29.7$k, with True vs False examples in the ratio of 1:10.
The skewed number of positive and negative labeled data reflects the real-life scenario where only a small portion of news are relevant to the target story 
(e.g., for the GE16 story, we found around $0.9$k relevant articles from a stream of $13$k news data).
Finally, we pair the 15 story representations to each of the documents. 
Thus, we have $445.5$k ($15\times 29.7$k) training data representing the story-doc relationship, 
mixing different sizes of stories and different types of documents.

\subsection{Feature Evaluation}\label{sec:feature-evaluation}

\begin{table*}[]
\centering
\caption{Evaluating the classification performance of different combinations of relevance features. The full set of 14 features is described in Table \ref{table:features}.}
\label{table:story-features}
\begin{tabular}{|l|c|c|c|c|}
\hline
\multicolumn{1}{|c|}{Features} & Precision & Recall & F1 & Runtime(s)    \\ \hline
Story/Doc Alone (1-4)         & 0.163     & 0.595  & 0.256  & 34.36\\ \hline
Story/Doc+Tf-idf (1-4,9-10)      & 0.757     & 0.632  & 0.689 & 63.38\\ \hline
Text-Based (1-10)               & 0.816     & 0.710  & 0.759 & 67.35\\ \hline
Graph-Based (1-4,9-14)         & 0.904     & 0.778  & 0.836 & 71.35\\ \hline
All Features(1-14)        & \textbf{0.907}     & \textbf{0.780}  & \textbf{0.839} & 72.63\\ \hline

\end{tabular}
\end{table*}

\begin{figure*}[]
	\centering 
    \includegraphics[width=0.9\linewidth]{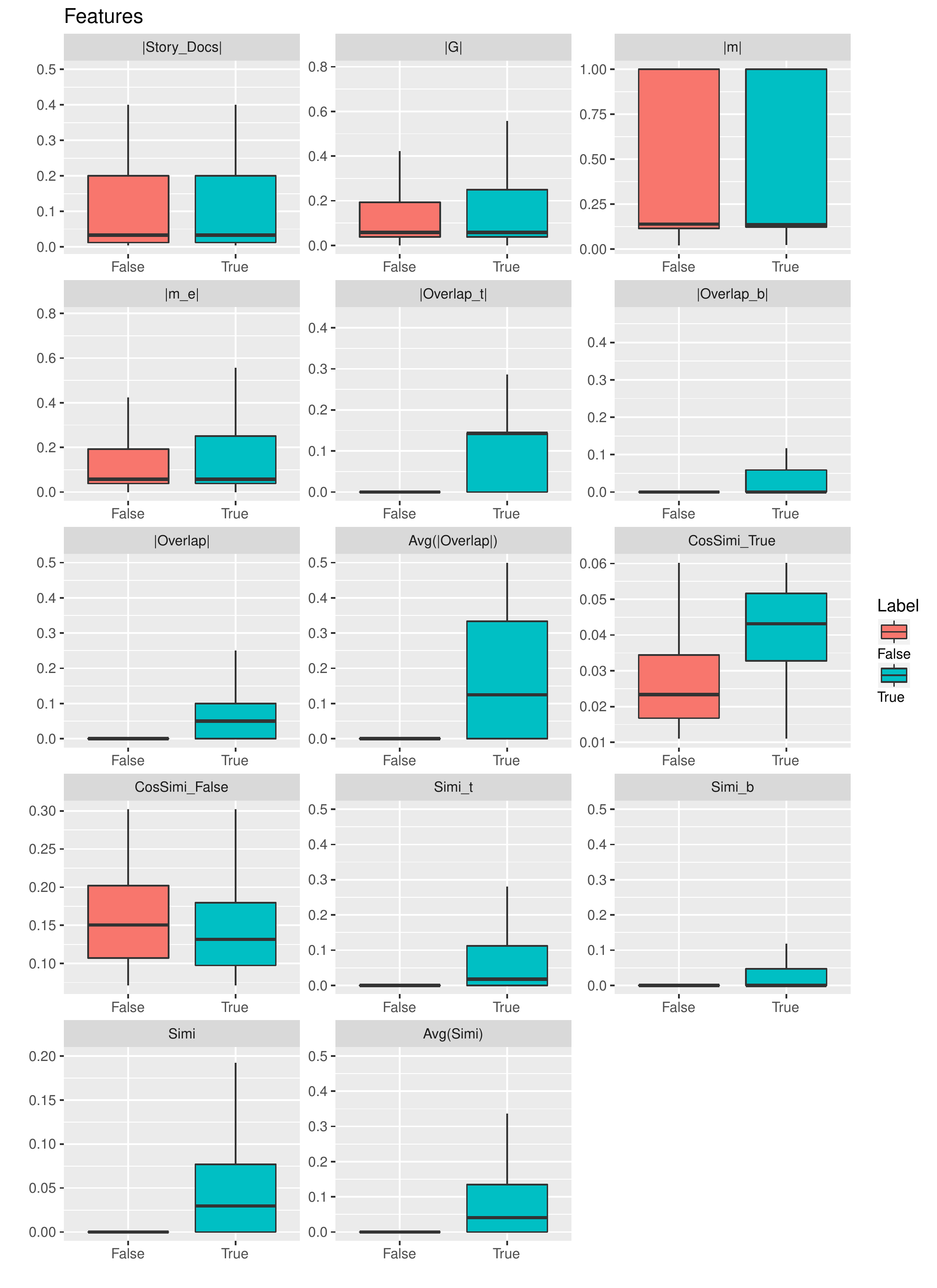}
%    \vspace{-0.4cm}
    \caption{The box plots of true and false labels of the 14 features used in the L2R model. Feature $|Overlap_{t}|$, $|Overlap|$, $Avg(|overlap|)$, $CosSimi\_True$, $Simi$ and $Avg(Simi)$ seem to be most useful for discriminating between relevant and irrelevant pairs, showing the impact of the entity-based, text-based, and graph-based features.}
    \label{fig:story-features}

\end{figure*}

In this section, we present a thorough analysis of the influence of different features on learning a story-doc relevance classifier.
Figure \ref{fig:story-features} shows the distribution in the 445.5k labeled data of the 14 features presented in Section \ref{sec:feature-engineering}.
The first 4 features only describe the story or the target document alone. Hence, we see similar distributions between the True and False labels.
The remaining features describe the association strength of a story and a document.
Feature $|Overlap_{t}|$, $|Overlap|$, $Avg(|overlap|)$, $CosSimi\_True$, $Simi$ and $Avg(Simi)$ seem to be most useful for discriminating between relevant and irrelevant pairs.
%The $Overlap$ features correspond to features 5-8 in Table \ref{table:features}, while $Simi$ features to features 11-14.

\textbf{Experiment Setup.} We train a RandomForest classifier on 445.5K labeled data, with a different combination of the 14 features.
The evaluation is done via 10-fold cross-validation.
We measure Precision, Recall, and F1 for evaluation, and record runtime for using different features in 10-fold cross-validation.

\textbf{Result.}
We test 5 combinations of the 14 features.
Experiment results are shown in Table \ref{table:story-features}.
Story/Document Alone (1-4) refers to the 4 features that only describe the story or the target document.
Since they do not capture the relevance between the story-doc pairs, using only such features gives poor classification results.
With two extra tf-idf features, which are used in most text mining techniques, we increase the F1 to 0.689.
Text-Based (1-10) adds the 4 $Overlap$ features presented in Table \ref{table:features}.
They are extracted from the disambiguated text content, and improve the classification quality by increasing the F1 to 0.759.
On the other hand, the Graph-Based (1-4,9-14) adds the 4 $Simi$ features that are extracted using the node weights of story graph $G$.
By comparing the Text-Based features to the Graph-Based, we demonstrate the importance of employing the story-entity graph representation to describe the story.
The classification quality increases considerably (F1 0.836 vs. 0.759).
We also notice that when using all 14 features, the Precision, Recall and F1 slightly increase comparing to the Graph-Based, showing that the Graph-Based features can replace the Text-Based features.
To ensure the robustness of the model in different datasets, we use all 14 features.

\subsection{Evaluating the Semi-Supervised Selection Methods}\label{sec:ssl-compare}

\begin{table*}[]
\centering
\caption{Comparing 5 semi-supervised selection (SSS) techniques using the Irish Water Charges story, when given 1 seed article. 
The $N$ and $E$ refer to number of nodes and edges of the initial story graph $G_{t_0}$. The N$\prime$ and E$\prime$ refer to the number of nodes and edges of the updated story graph when reaching the end of stream $T$. 
Time (seconds) refers to the running time of each method for the story and their corresponding test stream T, including both story updating time and document classification time.}
\label{table:sss-compare}
\begin{tabular}{|p{2cm}|c|c|c|c|c|c|c|c|c|}
\hline
Story                                        & N & E             & SSS    & N$\prime$ & E$\prime$ & Precision & Recall & F1   & Time(s) \\ \hline
\multirow{5}{2cm}{Irish Water Charges}         & \multirow{5}{*}{7} & \multirow{5}{*}{19} & None & 7              & 19             & 0.37      & 0.46   & 0.41 & 24   \\ \cline{4-10} 
                                             &                    &                     & Acc    & 2964            & 21124           & 0.30      & 0.74   & 0.42 & 601   \\ \cline{4-10} 
                                             &                    &                     & Rev    & 1858            & 13035           & 0.42      & 0.63   & 0.51 & 3507 \\ \cline{4-10} 
                                             &                    &                     & RR     & 2475            & 15531           & 0.38   & 0.56   & 0.45 &   2559   \\ \cline{4-10} 
                                             &                    &                     & \textbf{AR}     & 210            & 1042           & 0.38      & 0.71   & 0.49 &   101   \\ \hline
\end{tabular}
\end{table*}

In this section, we evaluate the 4 semi-supervised selection (SSS) techniques. 
In Section \ref{sec:semi-supervised}, we describe 2 standard SSS approaches from semi-supervised learning: Accumulate (Acc) and Revisit (Rev), and we propose 2 new SSS methods: Revisit-Recent (RR) and Accumulate+Revisit (AR).
We compare these 4 methods to using no SSS solution, on the Irish Water Charges story presented in Section \ref{sec:label-data}.
The Irish Water Charges is a long running story with articles and tweets gathered over a year.
During this period, the story developed quite a lot with a rich amount of details reported on all aspects of the story.

\textbf{Experiment Setup.} We simulate the story tracking process: given 1 article as \emph{seed doc} for a story, and a document stream $T$ (mix of articles and tweets), 
we classify the doc $m$, $m\in T$ into relevant or irrelevant to the story.
We first sort the labeled data by publishing time, then we use the first relevant article of the story as the \emph{seed doc}, and the rest as the stream $T$.
The number of relevant articles and tweets for Irish Water Charges are shown in Table \ref{table:9-stories-keyphrases-hashtags}. 
From the story stream, we sample 10 times more irrelevant articles and tweets as negative labeled data (1:10 positive to negative labeled data ratio in the seed).
We employ the L2R classifier trained with the 445.5k labeled data with all 14 features, and compare the performance of the 5 approaches using Precision, Recall, F1 and the total runtime.

\textbf{Result.} The experiment results are presented in Table \ref{table:sss-compare}. The 5 different SSS methods process the Irish Water Charges story. 
All methods use the same trained classifier, while the story representation (i.e., \emph{story docs} and entity graph $G$) are different between methods.
Because we only use a single article as the \emph{seed doc},
the first story entity graphs $G_{t_0}$ are very small (7 nodes and 19 edges).
At the end of the stream $T$, depending on the stories and the SSS methods, the story entity graphs are extended to tens or hundreds of nodes and edges (i.e., N$\prime$ and E$\prime$).

%The classification quality difference between the 5 SSS methods in the first 3 news stories is negligible.
%We find these 3 stories are straightforward to track, because of their low story complexity.
%These 3 stories have only one or two keyphrases that describe the story (e.g., cyclone debbie, kim jongnam), and almost all articles and tweets relevant to these story mentioned one of the keyphrases.
%The classification quality of these 3 stories are good enough (e.g., F1 around 90\%) even using no SSS method.

%Tracking the other 5 stories is more challenging. The EU Migrant Crisis, and 
%The Irish Water Charges is a long running stories with articles and tweets gathered over a year.
%During this period, the stories develop quite a lot with a rich amount of details reported on all aspects of the stories.
%The Trump Healthcare story is one of the many sub-stories of Trump and US politics, and thus, is very hard to separate this particular healthcare story from other Trump stories.
%The London Attack, on the other hand, is a short term, but a very popular story.
%A huge amount of articles and tweets are posted on this topic, and thus, include many story details.

With no SSS method and using only one article as \emph{seed doc}, the L2R classifier achieves around 0.41 in F1 when tracking Irish Water Charges story.
Among the 4 SSS solutions, Acc takes 601s, but in general, is not as accurate as the other 3 approaches.
Rev and RR have good accuracy, but take an extremely long time, making them not suitable for our real-time requirement.
The proposed AR approach achieves the best accuracy-time trade-off.
It only takes 101s, but increases the Precision, Recall and F1 by 2\%, 54\%, 19\%, a significant improvement compared to using no SSS solution.

\subsection{Comparing to Other Story Tracking Approaches}

In this experiment, we compare our model, Story Disambiguation, to state-of-the-art (SOTA) story tracking techniques.
There are three types of SOTA approaches: 1) binary classification methods that consider documents from the target story as one class and irrelevant documents as another; 2) story clustering/topic model methods that cluster documents into small groups and consider each group as a story/topic; 3) learning-to-rank approaches that evaluate the relevance of a story-document pair.
%From our initial experiment results, we find the performance of clustering techniques is not as good as the classification approaches, because they are designed to detect the first story from the document stream, rather than tracking stories.
%Hence, in this experiment, we only compare our model to binary classification approaches.
%These methods require some amount of labeled documents per story, and use bag-of-words type features.
We compare our proposed story disambiguation framework to 3 binary classification approaches, 2 cluster/topic modeling approaches and 1 L2R approach:
\begin{enumerate}
    \item \textbf{Text \cite{allan1998topic,schultz1999topic,makkonen2004simple}:} A binary classifier is trained on the positive and negative labeled examples for the target story, using tf-idf bag-of-words feature vectors. We compare some classifiers and choose Logistic Regression (in scikit-learn lib) for its accuracy and fast runtime when given large feature vectors.
    \item \textbf{Text+SSL \cite{Morinaga:2004:TDT:1014052.1016919,alsumait2008line}:} Similar to \textbf{Text},
    but adding a semi-supervised learning step that retrains the classifier based on past classification results. 
    The batch size is 50, and we use the Accumulation semi-supervised learning (SSL) approach in Section \ref{sec:ssl-compare}. This allows the classifier to adapt to the dynamic stream environment.
    \item \textbf{Text + Entity \cite{kumaran2005using}:} This method trains a binary classifier as the previous two approaches, but using both bag-of-words and bag-of-entities as the feature vector. The entities are extracted in the same style as in Section \ref{sec:ned}.
    \item \textbf{S-KMeans \cite{allan1998topic,yang1998study,brants2003system}:} A seeded K-Means algorithm is used to cluster the document streams into stories. 
    The initial cluster centroids are set using the training examples. The clusters that contain positive training examples are considered as relevant to the story.
    \item \textbf{DNMF \cite{takahashi2012applying}:} Dynamic topic modeling via non-negative matrix factorization (as detailed in \cite{greene2017exploring}). 
    The DNMF model extracts topics from the document streams for each time-window (i.e., one month) and finds coherent topics across time-windows. 
    The topics that contain positive training examples are considered as relevant to the story.
    \item \textbf{L2R \cite{lv2014pkuicst}:} A learning-to-rank classifier is trained on story-doc pairs of ``Irish General Election 2016'' story data (same as SD and SD+SSS), using tf-idf, BM25 and language models as similarity features. 
    The classifier is applied on the new story data.
      \end{enumerate}

We evaluate two versions of the Story Disambiguation framework:
\begin{enumerate}
    \item \textbf{Story Disambiguation (SD):} The standard Story Disambiguation framework without the semi-supervised selection (SSS) solution. SD uses a L2R ranking classifier trained on 445.5k labeled data gathered from ``Irish General Election 2016 (GE16)'' story. The positive training examples used by the SOTA approach are \emph{seed docs} in SD that represent the target story. The story entity graphs in this approach are static.
    \item \textbf{Story Disambiguation with Semi-Supervised Selection (SD+SSS):} The Story Disambiguation framework with SSS solution: Accumulate+Revisit (AR) as presented in Section \ref{sec:semi-supervised}. The story entity graphs in this approach are updated based on the predicted results of the L2R classifier.
\end{enumerate}

%As an evaluation metric, we use the standard Information Retrieval measurements: Precision, Recall, and F1. 

\subsubsection{Training Examples}

\begin{figure*}[]
    \centerline{
    \includegraphics[width=1.1\linewidth]{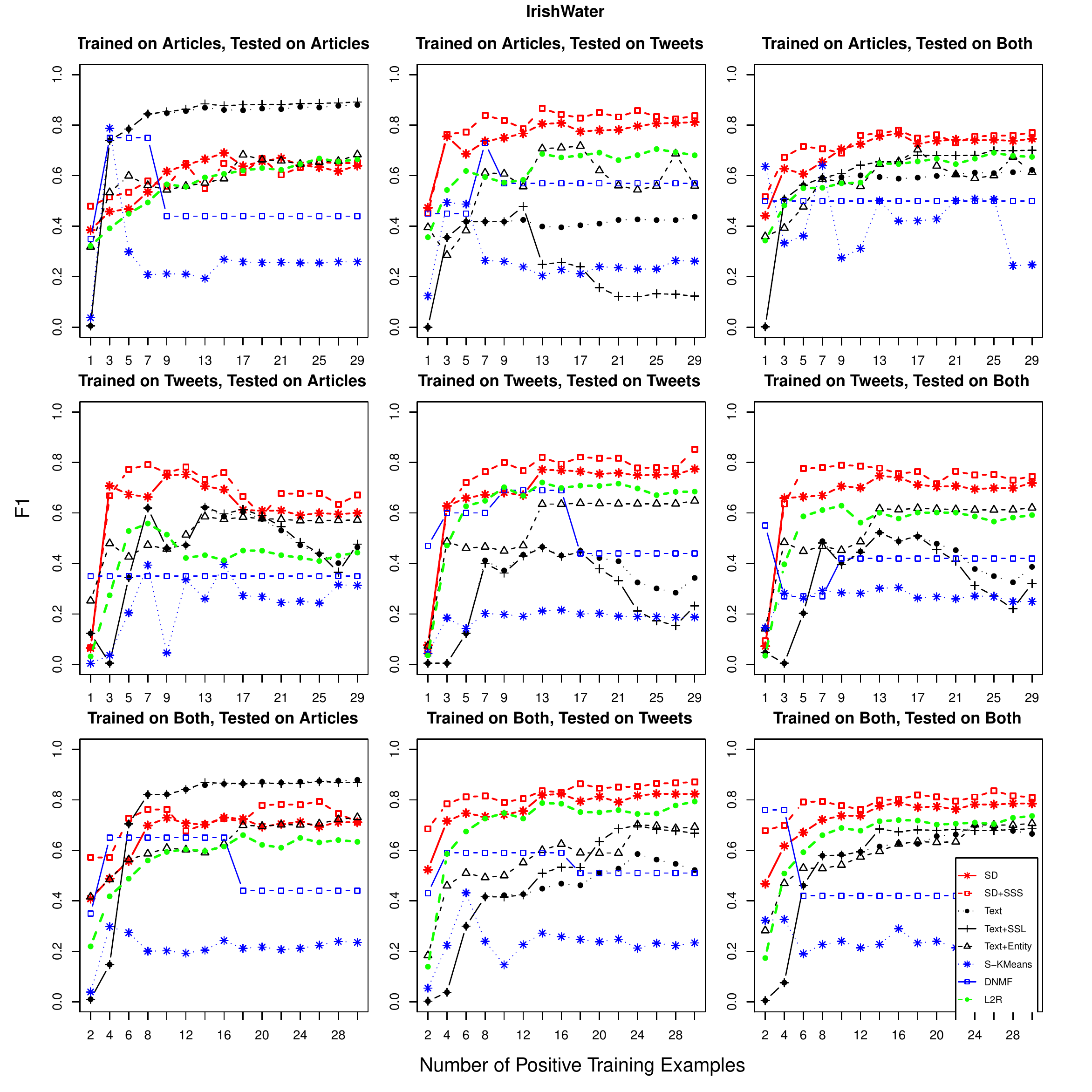}
    }
    \caption{Evaluating the F1 of the 8 compared methods given different number of positive training examples on the Irish Water Charges story. The type of training and testing data include: articles only, tweets only, and both articles and tweets. The experiment settings in the last row uses a minimum of 1 article and 1 tweet as seed documents.}
    \label{fig:compare-sota-train-text}
\end{figure*}

We first study the impact of a different number of training examples, as well as mixed types of training and testing examples (e.g., mixing articles and tweets).

\textbf{Experiment Setup.} We first evaluate the 8 compared methods on one story dataset, ``Irish Water Charges". 
The story dataset contains 389 relevant articles and 5k relevant tweets, and 10 times more irrelevant articles and tweets.
To analyse the behavior of these methods on data from different domains, we set-up 9 test cases where the training and test data are the combinations of three types: 
articles only, tweets only, and both articles and tweets.
For instance, one test case is to train the 8 methods on tweets and test them on articles.

In each test case, we vary the number of positive training examples for training the methods from 1 to 30 examples.
The number of negative training examples is 10 times that of the positive examples accordingly, 
and are selected randomly from the period close to the positive examples.
For test cases that use both articles and tweets as training examples, we select the same amount of articles and tweets.
We first sort the story dataset in chronological order, then we use the first 1-30 examples as the training data, while the rest as the testing data (i.e., simulating the document stream $T$).
We measure the F1 of the 8 methods.

\textbf{Result.} The detailed experiment results of the 9 test cases are presented in Figure \ref{fig:compare-sota-train-text}. 
The first row of the plot matrix is for test cases trained on articles, while the second row is for tweets and the last row is for both.
Similarly, the first column is for test cases that are tested on articles, and the other two columns are for testing on tweets, and both.
Each plot shows the F1 value of the 8 methods trained on a different number of training examples.

In terms of the number of training examples, in all 9 test cases, the binary classification approaches (Text, Text+SSL, and Text+Entity in black lines) need around 6 to 10 positive training examples to reach a stable F1.
On the other hand, the F1 of clustering approaches (S-KMeans, and DNMF in blue lines) is not stable and is generally better when given less than 10 positive labels.
The learning-to-rank approach (L2R in green line) also needs around 6 positive examples.
%The Test+Entity approach behave similarly, requiring around 5 positive examples, showing the impact of disambiguation techniques in story tracking tasks.
Our proposed Story Disambiguation methods SD and SD+SSS (in red lines), have good prediction quality (0.60 - 0.85 in F1) with 3-5 positive training examples.
Thus, we out-perform the other approaches when little labeled data is available.
This is due to the way we model the story tracking problem as an L2R problem and by representing the story in high-level entity graph.
By using a ranking classifier, we avoid retraining the model per target story, and thus, require less labeled data to define the story.

We have some interesting findings regarding the types of training and testing data.
First of all, the Text and Text+SSL approaches work better when trained and tested both on articles, as shown in subplot (1,1) (i.e., the plot on the first row and first column) in Figure \ref{fig:compare-sota-train-text}.
This shows the bag-of-words is a simple but powerful technique in the story tracking task.
On the other side, Text and Text+SSL do not work well when tested on tweets. 
%, comparing to Text+Entity. 
This is due to the bag-of-words features that cannot handle data of different language styles and vocabulary, 
while bag-of-entities is more robust in such cases (Text+Entity is a bit better in this case).

The clustering/topic model approaches have overall lower F1 than other methods regardless of the train/test data types.
This is because such techniques are designed to detect large and general topics instead of tracking specific stories.
The L2R approach has promising F1, but lower than our proposed methods, showing the power of the features in SD and SD+SSS.

%On the other side, if training and testing data are different (e.g., plot (1,2) and (2,1)), the SOTA approaches are not performing as well.
%In comparison, Entity and Text+Entity approaches perform much better, because they use the entities in the documents, which are disambiguated in advanced, and are very effective in describing the target story at a higher level.
SD and SD+SSS have stable prediction performance in F1 of 0.75 - 0.85, regardless of the train/test data types. 
These methods out-perform the others in most test cases, with SD+SSS better than the SD.
This shows our models are capable of tracking stories from multiple sources with different language style and vocabulary.

From the testing data perspective, regardless of the training data, articles are easier to track than tweets, with an average of 0.05 to 0.10 difference in F1 values.
Also, classifiers trained on articles are better in accuracy as compared to those trained on tweets, due to the short and noisy nature of the tweets.

\subsubsection{Target Stories}\label{sec:compare-stories}

In this section, we evaluate the 8 compared methods on different types of stories.

\textbf{Experiment Setup.}
We use the 8 test stories to compare the 8 methods trained and tested on both articles and tweets, same as in Figure \ref{fig:compare-sota-train-text} subplot (3,3).
This simulates the real-life scenario where the user provides a mixed set of documents as seed, and tracks stories in a mixed stream as well.
%Both cases use articles as training data, and both articles and tweets as testing data.
In this test case, the number of articles and tweets are equal, and the positive vs. negative examples ratio is 1:10.
Similar to the previous experiment, we sort the story datasets in chronological order, and use the first 1-15 articles and tweets as training data, and the rest of documents as test data.
We measure the F1 and running time of the 8 methods compared.
The amount of relevant articles and tweets for each story are shown in 
Table \ref{table:9-stories-keyphrases-hashtags}, and the attributes of the stories are given in Table \ref{table:9-stories}.

\begin{figure*}[]
    \centering 
    \includegraphics[width=0.9\linewidth]{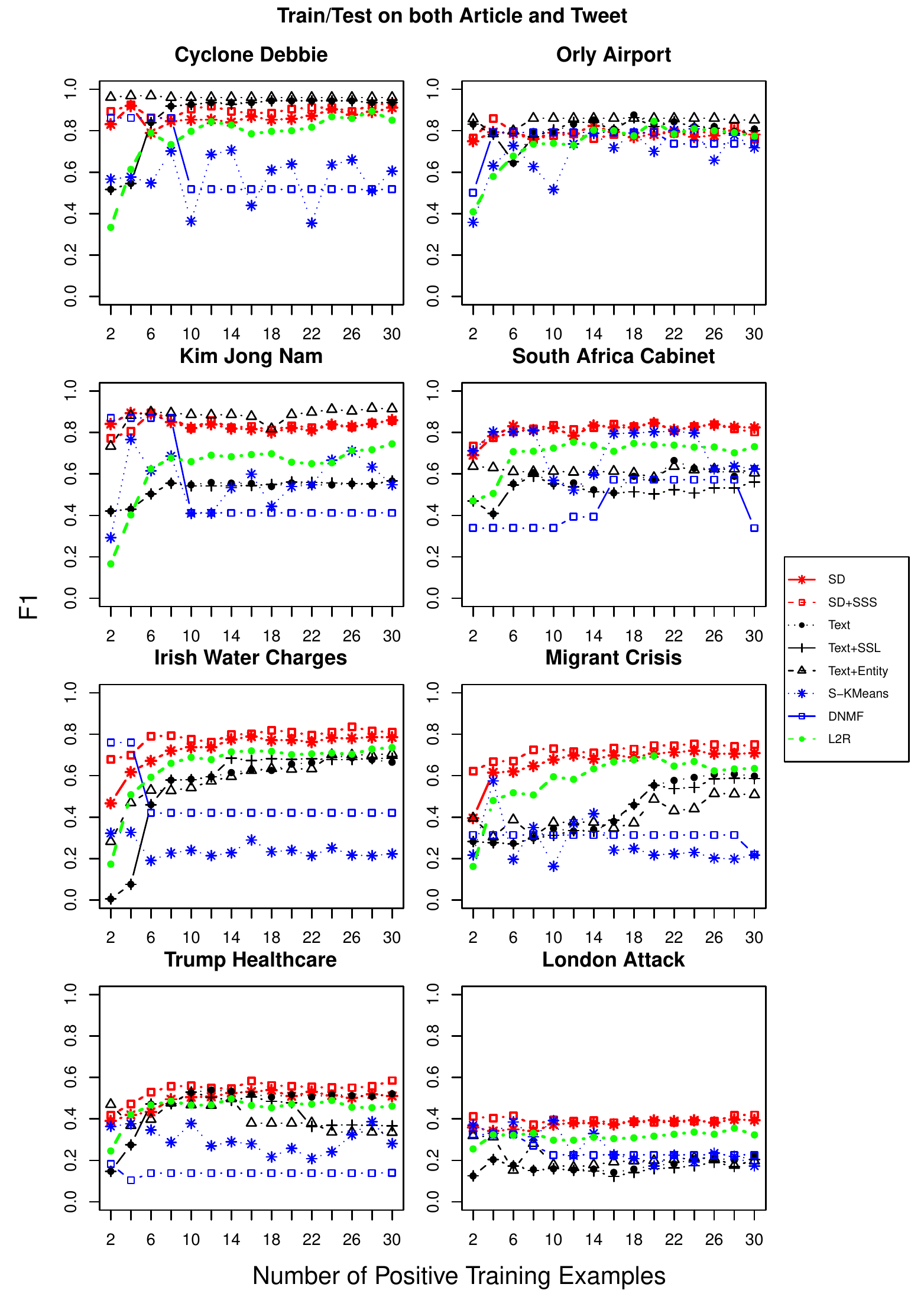}
    \caption{Evaluating the F1 of the 8 methods on 8 test stories given different number of training examples. The training and testing data are both articles and tweets. The number of articles and tweets in training examples are equal, and the positive negative examples ratio is 1:10.}
    \label{fig:compare-sota-8-stories}
\end{figure*}

\begin{figure}[]
    \centering 
    \includegraphics[width=1\linewidth]{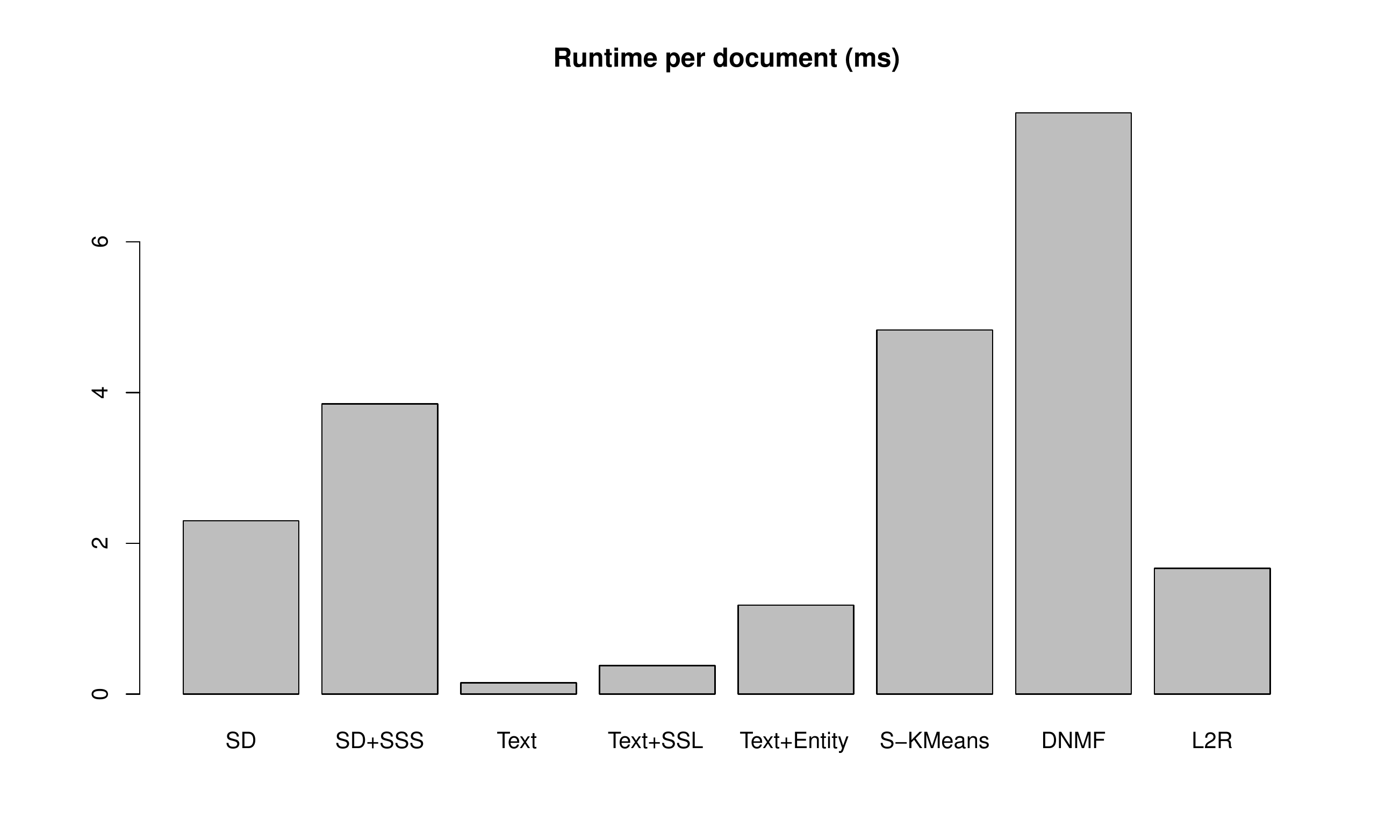}
    \caption{The runtime per document for the 8 methods compared.}
    \label{fig:sota-runtime}
\end{figure}

\textbf{Result.} The F1 of the 8 methods on the 8 test story datasets is shown in Figure \ref{fig:compare-sota-8-stories}.
%We also present the F1 values in Figure \ref{}.
%\todo{add a figure to present the results in table 5.8}
Across the 8 test stories, we see similar behaviors of the 8 methods as in the previous experiment.
Our proposed SD and SD+SSS have the best overall F1 and are stable regardless of the number of positive training examples, with the existing L2R method following closely. 
Text, Text+SSL, and Text+Entity approaches work well in low complexity stories, but F1 decreases quickly when the stories become challenging.
S-KMeans and DNMF are not as stable as other methods and generally perform worse.
%Our proposed SD and SD+SSS on the other hand, have stable performance regardless of the number of positive training examples.

%The 8 testing story are sorted by their complexity, and the stories at top are easier to tack comparing to the stories at bottom of the figure (F1 0.9 vs F1 0.4).

In terms of the 8 test stories, we find that the stories of high complexity are harder to track.
For low complexity stories, Cyclone Debbie, and Orly Airport Attack, the F1 of many methods easily goes beyond 0.85.
These stories are reported in the news with limited details and can be easily tracked with few story key phrases.
%The relevant documents are similar to each other, thus making these stories simple tracking targets.
The other 4 medium complexity stories are more difficult to track. 
The F1 of SOTA approaches drops quickly from 0.8 to 0.4 as the complexity of the story increases, while SD and SD+SSS are comparably stable.
%Given only 2 positive training example, the SOTA approaches are struggled with low F1. With enough training examples, the F1 of these approaches are mostly below 0.70.
For the two complex stories, all methods struggle, with the SD and SD+SSS slightly better than others.
%For instance, the F1 of Trump Healthcare story is below 0.40 for SOTA approaches, and F1 of London Attack is below 0.30.
%This is partially due to the insufficient training data. 
%For a complex story, binary classifier approaches requires more training examples to capture the story.
%This is due to that these stories have a complex structure between story entities, and the bag-of-words types of features are incapable of capturing such relationship.

In comparison to SOTA approaches, SD and SD+SSS achieve much better F1. 
Given the same number of training examples, their F1 is up to 0.20 higher than SOTA approaches in Trump Healthcare and London Attack story, 
up to 0.30 higher in the South Africa President Zuma, and EU Migrant Crisis story, and up to 0.50 higher in Irish Water Charges.
Also, their tracking quality is stable regardless of the number of training examples.
We believe this is due to our modeling and feature engineering.
By using an L2R classifier, we avoid retraining the model for each target story, and require less training examples to define the story.
Also, we capture the story entity structure in a high-level entity graph, where the PageRank algorithm weights the entity importance.
Thus, our model can better handle the story details in the complex stories.
In average, SD+SSS has higher F1 compared to SD, showing the impact of the semi-supervised selection techniques.

The runtime per documents for the 8 methods is shown in Figure \ref{fig:sota-runtime}. 
The binary classification approaches (Text, Text+SSL, and Text+Entity) take less amount of time: 0.15 ms, 0.38 ms, and 1.18 ms respectively.
The clustering approaches take the longest amount of time. Their high complexity makes them less suitable for real-time, large-scale streaming environments.
The SD, SD+SSS, and L2R all take a learning-to-rank approach, and their runtime is around 2 ms per documents. 
The runtime difference is due to the different features used in the models and the semi-supervised selection steps.

\section{Conclusion} 
\label{sec:story_conclusion}
%!TEX root = main.tex

In this paper, we have presented Story Disambiguation, a new framework for story tracking across multiple domains (e.g., news articles and tweets).
In contrast to state-of-the-art approaches that learn a classifier per target story, we model the story tracking task as a learning-to-rank problem, 
which allows us to train the classifier once, and re-use it for tracking new stories.
This significantly reduces the number of seed labeled data needed for defining a new story.
We employ entity disambiguation techniques and capture the story in a high-level entity graph, 
where we use biased PageRank to compute weights for entity importance.
By extracting features from both the entity ranking in the graph and the documents of the story, 
we are robust to the impact of the text language style, and are capable of tracking story content across multiple domains.
To deal with story drift over time, we propose new semi-supervised selection techniques for reusing unlabeled data in a streaming environment.

Our empirical study shows that our methods outperform the accuracy of the state-of-the-art when tracking stories across domains (e.g., news articles and tweets), 
and perform well on different types of stories, ranging from short, local stories, to complex, long-ranging stories.
Our proposed SD and SD+SSS approaches also need fewer positive examples to define the story and scale well as compared to the state-of-the-art methods, 
which makes them appealing for real-world story tracking application scenarios.
%while the SOTA approaches struggled in tracking complex stories that have a lot of story details, 
%as measured as the number of unique entities extracted from the story.

\section{Acknowledgement}
This work was supported by Science Foundation Ireland under Grant No. [SFI/12/RC/2289].

\end{document}